\documentclass[journal]{IEEEtai}

\usepackage{amssymb}
\usepackage{graphicx}
\usepackage{multirow}
\usepackage[position=top]{subfig}
\usepackage[table,xcdraw]{xcolor}
\usepackage[pagebackref,breaklinks,colorlinks,urlcolor=blue,linkcolor=blue,citecolor=blue]{hyperref}

\begin{document}

\title{Exploring Mutual Cross-Modal Attention for Context-Aware Human Affordance Generation}

\author{Prasun~Roy,~Saumik~Bhattacharya,~Subhankar~Ghosh,~Umapada~Pal,~and~Michael~Blumenstein
\vspace{-0.02em}
\thanks{Prasun~Roy and Subhankar~Ghosh and Michael~Blumenstein are with the University of Technology Sydney, NSW, Australia (e-mail: \{prasun.roy, subhankar.ghosh\}@student.uts.edu.au; michael.blumenstein@uts.edu.au).}
\thanks{Saumik~Bhattacharya is with the Indian Institute of Technology, Kharagpur, India (e-mail: saumik@ece.iitkgp.ac.in).}
\thanks{Umapada~Pal is with the Indian Statistical Institute, Kolkata, India (e-mail: umapada@isical.ac.in).}
\thanks{Code implementation is available at \url{https://github.com/prasunroy/mcma}.}
}

\markboth{January 2025}
{Exploring Mutual Cross-Modal Attention for Context-Aware Human Affordance Generation}

\maketitle

\begin{abstract}
Human affordance learning investigates contextually relevant novel pose prediction such that the estimated pose represents a valid human action within the scene. While the task is fundamental to machine perception and automated interactive navigation agents, the exponentially large number of probable pose and action variations make the problem challenging and non-trivial. However, the existing datasets and methods for human affordance prediction in 2D scenes are significantly limited in the literature. In this paper, we propose a novel cross-attention mechanism to encode the scene context for affordance prediction by mutually attending spatial feature maps from two different modalities. The proposed method is disentangled among individual subtasks to efficiently reduce the problem complexity. First, we sample a probable location for a person within the scene using a variational autoencoder (VAE) conditioned on the global scene context encoding. Next, we predict a potential pose template from a set of existing human pose candidates using a classifier on the local context encoding around the predicted location. In the subsequent steps, we use two VAEs to sample the scale and deformation parameters for the predicted pose template by conditioning on the local context and template class. Our experiments show significant improvements over the previous baseline of human affordance injection into complex 2D scenes.
\end{abstract}

\begin{IEEEImpStatement}
Human affordance generation is an intriguing yet challenging problem in computer vision. The ability to produce semantically coherent human instances in a complex scene can facilitate many downstream applications, including but not limited to synthetic data generation, augmented and virtual reality systems, digital media, and scene understanding. Existing methods have approached the problem using different network design philosophies without significant emphasis on contextual representation. This paper explores a novel cross-attention strategy for better semantic encoding in a disentangled human affordance generation scheme. Experimental studies show that the proposed method outperforms the existing techniques in qualitative and quantitative benchmarks.
\end{IEEEImpStatement}


\section{Introduction}
\label{sec:introduction}

\begin{figure*}[t]
  \centering
  \includegraphics[width=\linewidth]{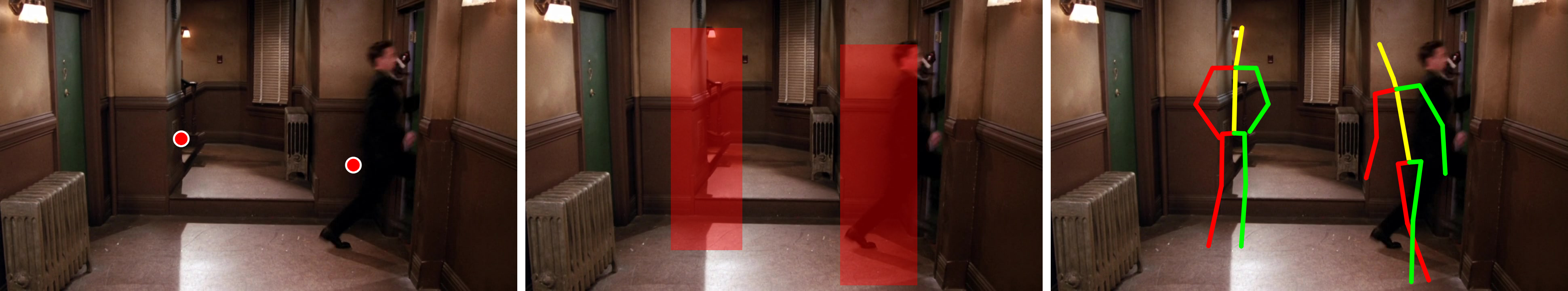}
  \caption{Overview of the proposed method. \textbf{Left:} Predicted locations for a new person in the scene. \textbf{Middle:} Estimated scale at each predicted location. \textbf{Right:} Final human pose estimated after scaling and deformation at each predicted location.}
  \label{fig:introduction}
\end{figure*}

The original conception \cite{gibson1979ecological} of relating perception with action describes affordance as the opportunities for interaction with the environment. In computer vision, human affordance prediction involves probabilistic modeling of novel human actions such that the estimated pose interprets a semantically meaningful interaction with the environment. The task is fundamental to many vision problems, such as machine perception, robot navigation, scene understanding, contextually sound novel human pose generation, and content creation. However, predicting a contextually relevant valid pose for a non-existent person is extremely challenging because, unlike the generic pose estimation task, we do not have an actual human body for supervision. So, in this case, the generator has to rely exclusively on the environmental semantics, requiring an intricate focus on the scene context representation.

Earlier works on affordance-aware human pose generation have explored knowledge base representation \cite{zhu2014reasoning}, social reasoning constraints \cite{chuang2018learning}, variational autoencoder \cite{wang2017binge}, adversarial learning \cite{wang2021scene, zhang2022inpaint} and transformer \cite{yao2023scene}. However, the majority of existing methods investigate different network design philosophies while putting less emphasis on the contextual representation of the scene. Unlike 3D, the lack of rich information about the surrounding environment in a 2D scene makes the problem particularly challenging and requires a robust representation of the scene context.
In this paper, we propose an affordance-aware human pose generation method in complex 2D indoor scenes by introducing a novel context representation technique leveraging cross-attention between two spatial modalities. The key idea is to mutually attend the convolution feature spaces from the scene image and the corresponding semantic segmentation map to get a modulated encoding of the scene context. Initially, we estimate a probable location where a person can be centered using a VAE conditioned on the global context encoding of the entire scene. After determining a potential center, local context embeddings are computed from square patches centered around that location at multiple scales. Next, we use a classifier on the multi-scale context vectors to predict the most likely pose as the template class from a set of existing human pose candidates. Finally, we use two VAEs conditioned on the context embeddings and the template class to sample the scale and linear deformations separately. The estimated scale and deformations are applied to the predicted pose template to get the target pose.

\vspace{1.20em}

\noindent
\textbf{Contributions:} The main contributions of the proposed work are summarized as follows.

\begin{itemize}
  \item We propose an affordance-aware human pose generation method in complex 2D indoor scenes by introducing a novel scene context representation technique that leverages mutual cross-attention between two spatial modalities for robust semantic encoding.
  \item Unlike the existing methods where the target position is user-defined, the proposed method utilizes global scene context to sample probable locations for the target, which provides additional flexibility for constructing a fully automated human affordance generation pipeline.
\end{itemize}

The remainder of the paper is organized as follows. We discuss the relevant literature in Sec. \ref{sec:related_work}. The proposed approach is discussed in Sec. \ref{sec:method}. Sec. \ref{sec:experiments} describes the dataset, experimental protocols, evaluation metrics, qualitative results, quantitative analysis, and a detailed ablation study. Finally, we conclude the paper in Sec. \ref{sec:conclusions} with a summary of key findings, potential use cases, and future scopes.

\section{Related Work}
\label{sec:related_work}

The original investigation \cite{gibson1979ecological} on the relationship between visual perception and human action defines \emph{affordance} as the opportunities for interaction with the surrounding environment. Behavioral studies on regular and cognitively impaired persons have shown evidence that perception results in both visual and motor signals in the human brain. An extended study \cite{anderson2002attentional} shows that visual attention to the spatial characteristics of the perceived objects initiates automatic motor signals for different actions. In computer vision, human affordance learning involves novel pose prediction such that the estimated pose represents a valid human action within the scene context. The task is fundamental to many problems requiring robust semantic reasoning about the environment, such as human motion synthesis \cite{wang2021scene} and scene-aware human pose generation \cite{wang2017binge, roy2016multi, zhang2022inpaint, yao2023scene}.

Earlier methods of affordance learning have explored knowledge mining \cite{zhu2014reasoning} and multimodal feature cues \cite{roy2016multi} to address the problem. In \cite{zhu2014reasoning}, the authors use a Markov Logic Network for constructing a knowledge base by extracting several object attributes from different image and metadata sources, which can perform various downstream visual inference tasks without any additional classifier, including zero-shot affordance prediction. In \cite{roy2016multi}, the authors use depth map, surface normals, and segmentation map as multimodal cues to train a multi-scale convolutional neural network (CNN) for scene-level semantic label assignment associated with specific human actions. In \cite{do2018affordancenet}, the authors design a multi-branch end-to-end CNN with two separate pathways for object detection and affordance label assignment to achieve high real-time inference throughput. Researchers \cite{chuang2018learning} have also explored socially imposed constraints for affordance learning. In \cite{chuang2018learning}, the authors propose a graph neural network (GNN) to propagate contextual scene information from egocentric views for action-object affordance reasoning.

Probabilistic modeling of scene-aware human motion generation also involves semantic reasoning of human interaction with the environment. Initial works on human motion synthesis have taken different architectural approaches, such as sequence-to-sequence models \cite{barsoum2018hp}, generative adversarial networks (GAN) \cite{barsoum2018hp, cai2018deep, yang2018pose}, graph convolutional networks (GCN) \cite{yan2019convolutional}, and variational autoencoders (VAE) \cite{guo2020action2motion}. However, these methods have mostly ignored the role of environmental semantics. Due to potential uncertainty in human motion, in a recent approach \cite{wang2021scene}, the authors address such motion synthesis with a GAN conditioned on scene attributes and motion trajectory to predict probable body pose dynamics.

One key challenge of human affordance generation in 2D scenes is the lack of large-scale datasets with rich pose annotations. In \cite{wang2017binge}, the authors compile the only public dataset of annotated human body poses in complex 2D indoor scenes by extracting frames from sitcom videos. Aiming to generate a contextually valid human affordance at a user-defined location, the authors propose sampling the scale and deformation parameters for an existing human pose template using a VAE conditioned on the localized image patches as scene context. In \cite{zhang2022inpaint}, the authors introduce a two-stage GAN architecture for achieving a similar goal by estimating the affine bounding box parameters to localize a probable human in the scene and then generating a potential body pose at that location. The method uses the input scene, corresponding depth, and segmentation maps as semantic guidance. In \cite{yao2023scene}, the authors propose a transformer-based approach with knowledge distillation for generating human affordances in 2D indoor scenes.

\section{Method}
\label{sec:method}

Generating a semantically meaningful body pose of a non-existent person is challenging. Because, unlike a traditional pose estimation approach, an actual human is not present for supervision of the body joints, which forces the generator to depend exclusively on the scene context. However, directly using the spatial feature maps from one \cite{wang2017binge, do2018affordancenet, yao2023scene} or multiple \cite{chuang2018learning, wang2021scene, zhang2022inpaint} modalities does not provide a comprehensive encoding of the scene context, thereby limiting the sampling quality of the generator. Our key idea is a bidirectional cross-attention mechanism between feature spaces of two modalities for a modulated scene context representation. In particular, we use an ImageNet-pretrained \cite{deng2009imagenet} VGG-19 model \cite{simonyan2015very} to estimate the convolution feature maps from the scene image and the corresponding segmentation mask, followed by mutually cross-attending the two feature spaces. The proposed affordance generation pipeline consists of four stages. In the first stage, we use a conditional VAE to estimate a probable location within the scene where a person can be centered. In the second stage, a classifier predicts the most likely template pose for the estimated location from a set of existing human pose candidates. In the subsequent stages, we use two conditional VAEs to sample the scale and linear deformation parameters for the predicted template. Fig. \ref{fig:architecture} illustrates an overview of the proposed architecture.

\subsection{Context representation}
\label{sec:method_context}

\begin{figure}[t]
  \centering
  \includegraphics[width=\linewidth]{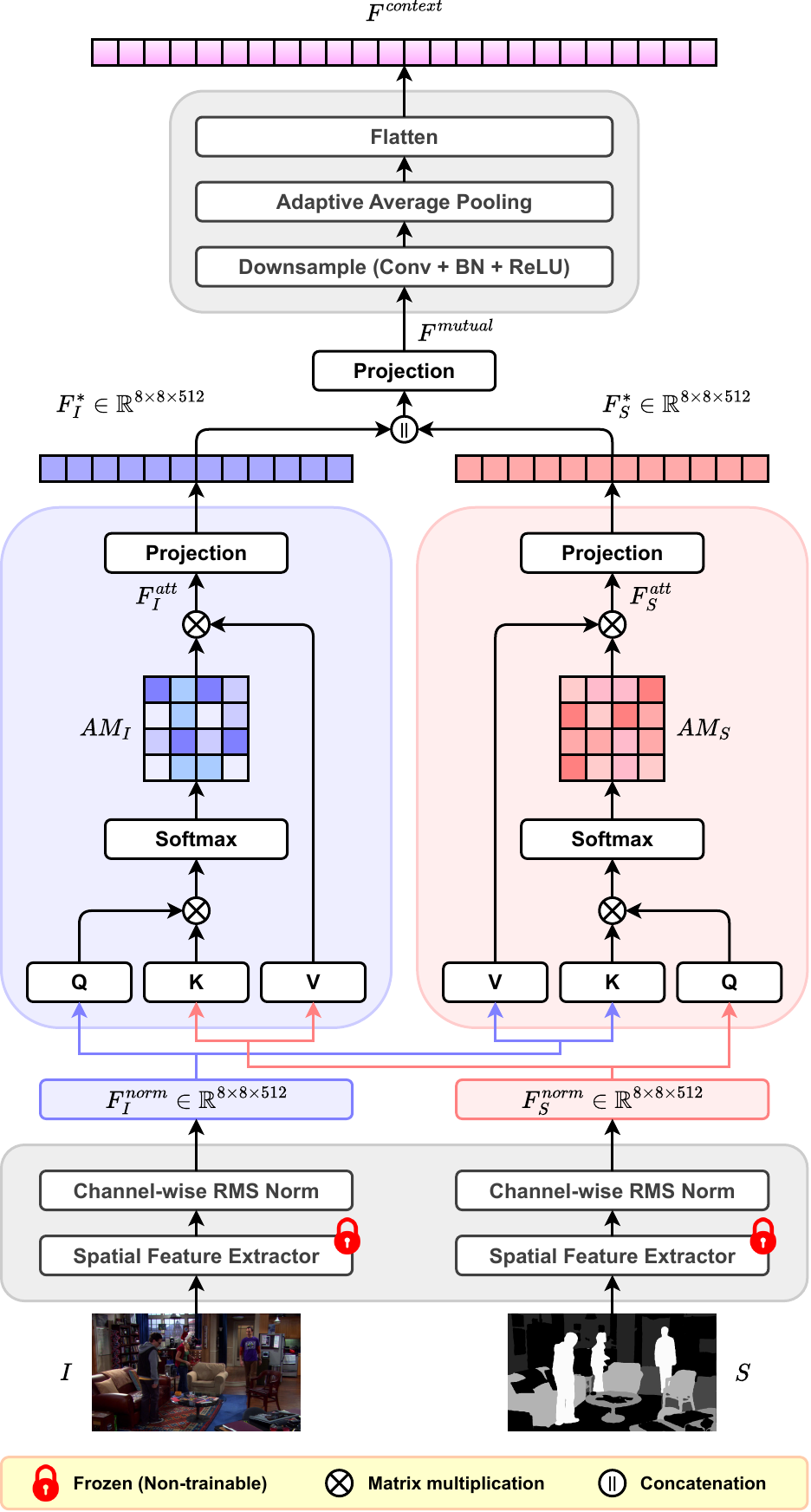}
  \caption{Architecture of the \emph{Mutual Cross-Modal Attention} (\textbf{MCMA}) block.}
  \label{fig:mcma_block}
  \vspace{-1.0em}
\end{figure}

Conceptually, the attention mechanism \cite{vaswani2017attention} dynamically increases the receptive field in a network architecture \cite{wang2018non}. Unlike self-attention, which combines two similar embedding spaces, cross-attention asymmetrically combines two separate embedding spaces. More specifically, self-attention computes the attention maps from queries $Q$, keys $K$, and values $V$ projected from the same token space, while cross-attention uses $K$ and $V$ projected from a different token space as the contextual guidance. However, the underlying computation steps are identical in both cases and involve a similarity measure between $Q$ and $K$ to form an attention matrix $AM$, followed by weighing $V$ with $AM$ to obtain the updated query tokens.

In the proposed cross-attention mechanism, we use an image $I$ and the corresponding semantic segmentation map $S$ as inputs from two different modalities for mutually attending to each other. The segmentation maps are estimated from respective images using OneFormer \cite{jain2023oneformer} with DiNAT-L \cite{hassani2022dilated} backbone pretrained on the ADE20K dataset \cite{zhou2019semantic}. We reduce the initial 150 semantic labels to 8 most significant categories to represent indoor scenes as grayscale semantic maps with the following values -- wall (36), floor (72), stairs (108), table (144), chair (180), bed (216), person (252), and background/everything else (0). Initially, we resize $I$ and $S$ with a rescaling function $f: \mathbb{R}^{h \times w \times 3} \rightarrow \mathbb{R}^{256 \times 256 \times 3}$. The resized images are passed through the convolutional backbone of an ImageNet-pretrained \cite{deng2009imagenet} frozen VGG-19 model \cite{simonyan2015very} to extract the respective spatial feature maps $F^{conv}_I$ and $F^{conv}_S$, $g: \mathbb{R}^{256 \times 256 \times 3} \rightarrow \mathbb{R}^{8 \times 8 \times 512}$. The resulting feature maps $F^{conv}_I$ and $F^{conv}_S$ are normalized with channel-wise root mean square layer normalization (RMSNorm) \cite{zhang2019root} to obtain the normalized feature maps $F^{norm}_I$ and $F^{norm}_S$. Mathematically, the normalization operation defines a function $\phi: \mathbb{R}^{8 \times 8 \times 512} \rightarrow \mathbb{R}^{8 \times 8 \times 512}$ as
\[
F^{norm} = \phi(F^{conv}) = \gamma \; \frac{F^{conv} \sqrt{N_c}}{max(\|F^{conv}\|_2, \epsilon)}
\]
where $N_c$ denotes the number of channels in feature map $F^{conv}$, $\epsilon \in \mathbb{R}^n$ is a very small constant $(\approx 1e^{-12})$ to provide numerical stability, and $\gamma \in \mathbb{R}^n$ is a learnable gain parameter set to 1 at the beginning.

For computing the multi-head attention maps, we first define three projection functions $q$, $k$, and $v$ to map the normalized feature maps $F^{norm}$ into queries $Q$, keys $K$, and values $V$ as
\[
Q = q(F^{norm}): \mathbb{R}^{8 \times 8 \times 512} \rightarrow \mathbb{R}^{8 \times 8 \times N_{embed}}
\]
\[
K = k(F^{norm}): \mathbb{R}^{8 \times 8 \times 512} \rightarrow \mathbb{R}^{8 \times 8 \times N_{embed}}
\]
\[
V = v(F^{norm}): \mathbb{R}^{8 \times 8 \times 512} \rightarrow \mathbb{R}^{8 \times 8 \times N_{embed}}
\]
where the projection functions use point-wise convolution with $1 \times 1$ kernels and zero bias, and $N_{embed}$ is the length of the cumulative embedding space of all attention heads. In the proposed architecture, we use 8 attention heads of dimension 64 each, resulting $N_{embed} = 8 \times 64 = 512$. Next, we update the feature maps from both modalities $I$ and $S$ by computing the respective cross-attention matrices, taking $Q$ from one modality and $K$ \& $V$ from the other. Mathematically,
\[
F^{att}_I = softmax \left( \frac{Q_I K^T_S}{\sqrt{d}} \right) V_S
\]
\[
F^{att}_S = softmax \left( \frac{Q_S K^T_I}{\sqrt{d}} \right) V_I
\]
where $d$ is a scaling parameter. In the proposed method, we take the attention head dimension as $d$.

The final updated cross-modal feature maps are estimated as $F^*_I = p(F^{att}_I)$ and $F^*_S = p(F^{att}_S)$, where the function $p: \mathbb{R}^{8 \times 8 \times N_{embed}} \rightarrow \mathbb{R}^{8 \times 8 \times 512}$ defines a projection that uses convolution with $1 \times 1$ kernels and zero bias.

Finally, we compute the mutual cross-modal feature maps as $F^{mutual} = t(c(F^*_I, F^*_S))$ by a channel-wise concatenation $c: \mathbb{R}^{8 \times 8 \times 512} \times \mathbb{R}^{8 \times 8 \times 512} \rightarrow \mathbb{R}^{8 \times 8 \times 1024}$ of $F^*_I$ and $F^*_S$, followed by a projection $t: \mathbb{R}^{8 \times 8 \times 1024} \rightarrow \mathbb{R}^{8 \times 8 \times 512}$ using convolution with $1 \times 1$ kernels and zero bias.

The vectorized representation of the contextual embedding $F^{context} \in \mathbb{R}^{8192}$ is obtained by first downsampling the $8 \times 8$ feature maps of $F^{mutual}$ into a spatial resolution of $4 \times 4$ with strided convolution ($4 \times 4$ kernel, stride = 2, padding = 1, bias = 0), then following a set of sequential operations -- batch normalization, ReLU activation, $4 \times 4$ adaptive average pooling, and flattening. We use $F^{context}$ as the semantic condition in the proposed architecture. The architecture of the proposed \emph{mutual cross-modal attention} (\textbf{MCMA}) block is illustrated in Fig. \ref{fig:mcma_block}.

\subsection{Estimating locations of non-existent persons}
\label{sec:method_location}
Unlike the existing approaches, where the location of the target person is user-specified, we attempt to estimate the probable locations of non-existent persons automatically to minimize user intervention. However, the problem is challenging as potential human candidates may appear at several locations with varying scales and poses. Therefore, inferring a location directly from spatial feature space is difficult. In the proposed method, we use a VAE to sample a possible location by conditioning the network on the global context embedding $F^{context} \in \mathbb{R}^{2048}$, computed over the entire scene. The encoder and decoder in the VAE architecture use a shared feature space $F^{shared} \in \mathbb{R}^{128}$, obtained by projecting $F^{context}$ into a 128-dimensional vector with a fully connected (FC) layer and ReLU activation.

In the encoder network, the 2D location coordinates $o(x, y) \in \mathbb{R}^2$ is first mapped into a 128-dimensional vector using two consecutive FC layers having ReLU activations, and the output is linearly concatenated with $F^{shared}$. We compute the mean $\mu \in \mathbb{R}^{32}$ and variance $\sigma \in \mathbb{R}^{32}$ of the latent distribution $P(\mu, \sigma)$ by projecting the concatenated feature vector through two separate FC layers. The latent embedding vector $z$ is obtained using the reparameterization technique \cite{kingma2013auto} as $z = \mu + \sigma \odot \epsilon$, where $z \in \mathbb{R}^{32}$ is sampled from the estimated distribution $P(z \; | \; \mu, \sigma)$ and $\epsilon \in \mathbb{R}^{32}$ is sampled from the normal distribution $\mathcal{N}(0, 1)$.

In the decoder network, the latent embedding $z$ is projected into a 128-dimensional space using two consecutive FC-ReLU layers. The shared feature vector $F^{shared}$ also receives an additional projection into a 128-dimensional space through a single FC-ReLU layer. The two projected vectors are linearly concatenated and passed over another 128-dimensional FC-ReLU layer. Finally, we use one more FC layer to project the feature space into the predicted 2D location coordinates $\overline{o}(x, y) \in \mathbb{R}^2$.

The optimization objective for the network consists of two loss components. To measure the spatial deviation, we compute the mean squared error ($\mathcal{L}_{MSE}$) between the 2D target and predicted locations $o(x, y)$ \& $\overline{o}(x, y)$ as follows.
\[
\mathcal{L}_{MSE} = \| \; \overline{o}(x, y) \; - \; o(x, y) \; \|_2
\]
To evaluate the statistical difference between the estimated probability distribution of the embedding space $P(z \; | \; \mu, \sigma)$ and the normal distribution $\mathcal{N}(0, 1)$, we compute the Kullback–Leibler divergence \cite{kullback1951information} ($\mathcal{L}_{KLD}$) between the two distributions as follows.
\[
\mathcal{L}_{KLD} = KL \left( P(z \; | \; \mu, \sigma) \; \| \; \mathcal{N}(0, 1) \right)
\]

We update the network parameters by minimizing the cumulative objective $\mathcal{L} = \mathcal{L}_{MSE} + \mathcal{L}_{KLD}$ using stochastic Adam optimizer \cite{kingma2015adam}, keeping a fixed learning rate of $1e^{-3}$ and $\beta$-coefficients at (0.5, 0.999). During inference, we estimate a probable center $o^*(x, y)$ for a non-existent person by using a random noise $\eta \in \mathbb{R}^{32}$, sampled from the normal distribution $\mathcal{N}(0, 1)$, and the shared scene context embedding $F^{shared} \in \mathbb{R}^{128}$ as inputs to the VAE decoder $\mathcal{D}^{vae}$. Formally, $o^*(x, y) = \mathcal{D}^{vae}(\eta, F^{shared})$, $\eta \sim \mathcal{N}(0, 1)$.

\subsection{Finding pose templates for non-existent persons}
\label{sec:method_pose}
Unlike the conventional human pose estimation techniques \cite{cao2017realtime, sun2019deep, xu2022vitpose, geng2023human}, directly inferring the valid pose of a non-existent person is difficult \cite{roy2022scene} due to the unavailability of an actual human body for supervision. Thus, after sampling a probable location $o^*(x, y)$ within the scene where a person can be centered, we select a potential candidate from an existing set of $m$ valid human poses as the initial guess (template) at that position. In the next stage, the template pose is scaled and deformed to estimate the target pose. The candidate pool is constructed using the K-medoids algorithm \cite{park2009simple} to select $m$ representative pose templates from all the available human poses in the training data. Each template class is represented as a $m$-dimensional one-hot embedding vector $y \in \mathbb{R}^m$. To estimate the class probabilities $\overline{y} \in \mathbb{R}^m$ for an expected pose, we use a multi-class classifier on the context embedding.

The global cross-modal feature maps $F^{mutual} \in \mathbb{R}^{8 \times 8 \times 512}$ are estimated over the entire scene as discussed in Sec. \ref{sec:method_context}. In addition to the global features, we compute two local feature representations $F^{mutual}_A \in \mathbb{R}^{8 \times 8 \times 512}$ and $F^{mutual}_B \in \mathbb{R}^{8 \times 8 \times 512}$ over two square patches $A$ and $B$, centered at the initially sampled location $o^*(x, y)$. The size of patch $A$ is equal to the scene height, and the size of patch $B$ is half of the scene height. A channel-wise concatenation of the global and local feature maps represents the cumulative feature space $F^{mutual}_* \in \mathbb{R}^{8 \times 8 \times 1536}$. We derive the combined context embedding vector $F^{context}_* \in \mathbb{R}^{8192}$ by first downsampling $F^{mutual}_*$ into a spatial dimension of $\mathbb{R}^{4 \times 4 \times 512}$ with strided convolution ($4 \times 4$ kernel, stride = 2, padding = 1, bias = 0), then following a set of sequential operations similar to Sec. \ref{sec:method_context} -- batch normalization, ReLU activation, $4 \times 4$ adaptive average pooling, and flattening. Finally, the classifier predicts the multi-class probabilities by projecting the context embedding $F^{context}_*$ into a $m$-dimensional output vector $\overline{y} \in \mathbb{R}^m$ using an FC-layer, followed by $softmax$ activation.

The optimization objective of the classifier is a multi-class (categorical) cross-entropy loss ($\mathcal{L}_{CCE}$), which is formally defined using the negative $log$-likelihood as follows.
\[
\mathcal{L}_{CCE} = - \sum_{i=1}^m y_i \; log(\overline{y}_i)
\]
where $y_i \in y$ and $\overline{y}_i \in \overline{y}$ denote the probabilities of $i$-th class, $i \in \{1, ..., m\}$, in the target ($y$) and predicted ($\overline{y}$) label vectors, respectively. We update the network parameters by minimizing $\mathcal{L}_{CCE}$ using stochastic Adam optimizer \cite{kingma2015adam}, keeping a fixed learning rate of $1e^{-3}$ and $\beta$-coefficients at (0.5, 0.999). During inference, we obtain the one-hot pose template class embedding $y^* \in \mathbb{R}^m$ by selecting the predicted class with the highest probability. Formally, $y^* = argmax(\overline{y})$.

\subsection{Scaling the selected pose template}
\label{sec:method_scale}
After inferring a probable location $o^*(x, y)$ and one-hot pose template class embedding $y^* \in \mathbb{R}^m$, we use a conditional VAE to sample the expected scaling factors (height and width) of the target person. The estimated parameters are used to rescale the normalized pose template of unit height and width into the target dimensions. The encoder and decoder networks of the VAE use a shared feature space $F^{shared}_* \in \mathbb{R}^{128}$ as the condition, which is derived from the cumulative context vector $F^{context}_* \in \mathbb{R}^{2048}$ and pose template class embedding $y^* \in \mathbb{R}^m$. As discussed in Sec. \ref{sec:method_pose}, $F^{context}_*$ encodes both global (over entire scene) and local (over localized patches) cross-modal scene context representations. To compute the shared feature space, we first project $y^*$ into a 128-dimensional vector by two consecutive FC-ReLU layers and then linearly concatenate the projected output with $F^{context}_*$. The concatenated vector is passed through another FC-ReLU layer to obtain the 128-dimensional shared feature representation $F^{shared}_* $.

The encoder and decoder of the VAE adopt a similar architecture as the location estimator discussed in Sec. \ref{sec:method_location}. The encoder takes the scaling parameters $s(\Delta x, \Delta y) \in \mathbb{R}^2$ and shared feature vector $F^{shared}_* \in \mathbb{R}^{128}$ as inputs to predict the mean $\mu_s \in \mathbb{R}^{32}$ and variance $\sigma_s \in \mathbb{R}^{32}$ of the latent distribution $P_s(\mu_s, \sigma_s)$. The decoder takes the latent embedding $z_s \in \mathbb{R}^{32}$ and shared feature vector $F^{shared}_* \in \mathbb{R}^{128}$ as inputs to estimate the probable scaling parameters $\overline{s}(\Delta x, \Delta y) \in \mathbb{R}^2$, where $z_s$ is computed using the reparameterization technique as $z_s = \mu_s + \sigma_s \odot \epsilon$, $z_s \sim P_s(\mu_s, \sigma_s)$, and $\epsilon \sim \mathcal{N}(0, 1)$.

The optimization objective for the network consists of two loss components. The first term measures the spatial deviation $\mathcal{L}_{MSE}$ between $s(\Delta x, \Delta y)$ and $\overline{s}(\Delta x, \Delta y)$ as the $L_2$ norm. The second term estimates the statistical difference $\mathcal{L}_{KLD}$ between $P_s(z_s \; | \;\mu_s, \sigma_s)$ and $\mathcal{N}(0, 1)$ as the KL divergence. Mathematically, the loss terms can be represented as follows.
\[
\mathcal{L}_{MSE} = \| \; \overline{s}(\Delta x, \Delta y) \; - \; s(\Delta x, \Delta y) \; \|_2
\]
\[
\mathcal{L}_{KLD} = KL \left( P_s(z_s \; | \; \mu_s, \sigma_s) \; \| \; \mathcal{N}(0, 1) \right)
\]

We update the network parameters by minimizing the cumulative objective $\mathcal{L} = \mathcal{L}_{MSE} + \mathcal{L}_{KLD}$ using stochastic Adam optimizer, with a fixed learning rate of $1e^{-3}$ and $\beta$-coefficients of (0.5, 0.999). During inference, we estimate the probable scaling factors $s^*(\Delta x, \Delta y)$ for the selected pose template by using a random noise $\eta \in \mathbb{R}^{32}$, sampled from $\mathcal{N}(0, 1)$, and the shared scene context embedding $F^{shared}_* \in \mathbb{R}^{128}$ as inputs to the VAE decoder $\mathcal{D}^{vae}_s$. Formally, $s^*(\Delta x, \Delta y) = \mathcal{D}^{vae}_s(\eta, F^{shared}_*)$, $\eta \sim \mathcal{N}(0, 1)$.

\subsection{Deforming the selected pose template}
\label{sec:method_deform}
The potential human pose at the sampled location $o^*(x, y)$ is estimated by applying a linear deformation on the chosen pose template. A linear deformation vector $d$ is the set of distances between the cartesian coordinates of each body keypoint. Assuming a human pose is represented with $r$ keypoints, we define $d = \{dx_1, dy_1, ..., dx_r, dy_r\} \in \mathbb{R}^{2r}$, where ($dx_j, dy_j$) denotes the differences between the template and target coordinates along $x$ and $y$ axes for the $j$-th keypoint, $1 \leqslant j \leqslant r$. In the proposed method, we represent a human pose with 16 major body joints ($r = 16, d \in \mathbb{R}^{32}$), following the MPII \cite{andriluka142d} keypoint format.

The deformation parameters are estimated using an identical VAE architecture as discussed in Sec. \ref{sec:method_scale}. The encoder takes the context embedding $F^{shared}_* \in \mathbb{R}^{512}$ and linear deformations $d \in \mathbb{R}^{32}$ as inputs to predict the mean $\mu_d \in \mathbb{R}^{32}$ and variance $\sigma_d \in \mathbb{R}^{32}$ of the latent distribution $P_d(\mu_d, \sigma_d)$. The decoder takes $F^{shared}_* \in \mathbb{R}^{512}$ and the latent vector $z_d \in \mathbb{R}^{32}$ as inputs to estimate the probable deformation parameters $\overline{d} \in \mathbb{R}^{32}$, where $z_d = \mu_d + \sigma_d \odot \epsilon$, $z_d \sim P_d(\mu_d, \sigma_d)$, and $\epsilon \sim \mathcal{N}(0, 1)$.

The network parameters are optimized by minimizing the cumulative objective $\mathcal{L} = \mathcal{L}_{MSE} + \mathcal{L}_{KLD}$ using stochastic Adam optimizer, with a fixed learning rate of $1e^{-3}$ and $\beta$-coefficients of (0.5, 0.999). Mathematically,
\[
\mathcal{L}_{MSE} = \| \; \overline{d} \; - \; d \; \|_2
\]
\[
\mathcal{L}_{KLD} = KL \left( P_d(z_d \; | \; \mu_d, \sigma_d) \; \| \; \mathcal{N}(0, 1) \right)
\]

The probable deformation parameters $d^* \in \mathbb{R}^{32}$ for the selected pose template is inferred by using a random noise $\eta \in \mathbb{R}^{32}$ and the shared scene context embedding $F^{shared}_* \in \mathbb{R}^{512}$ as inputs to the VAE decoder $\mathcal{D}^{vae}_d$. Formally, $d^* = \mathcal{D}^{vae}_d(\eta, F^{shared}_*)$, $\eta \sim \mathcal{N}(0, 1)$.

\subsection{Target transformation}
\label{sec:method_transformation}

Assuming a normalized human pose template of unit scale $h^*(x_1, y_1, ..., x_r, y_r)$ corresponding to the predicted pose template class $y^*$, we compute the target pose $\overline{h}(\overline{x}_1, \overline{y}_1, ..., \overline{x}_r, \overline{y}_r)$ from the estimated center $o^*(x_0, y_0)$, scaling factors $s^*(\Delta x, \Delta y)$, and linear deformations $d^*(dx_1, dx_2, ..., dx_r, dy_r)$ as follows.
\[
\overline{x}_i = \frac{w}{w_0} \left[ \left( x_i \Delta x + dx_i \right) + \left( x_0 - \frac{\Delta x}{2} \right) \right]
\]
\[
\overline{y}_i = \frac{h}{h_0} \left[ \left( y_i \Delta y + dy_i \right) + \left( y_0 - \frac{\Delta y}{2} \right) \right]
\]
where $i \in \{1, 2, ..., r\}$, $(h_0, w_0)$ is the rescaled image patch size for network input, and $(h, w)$ is the size of the original scene. In the proposed method, we use $h_0 = w_0 = 256$.

\begin{figure*}[t]
  \centering
  \includegraphics[width=\linewidth]{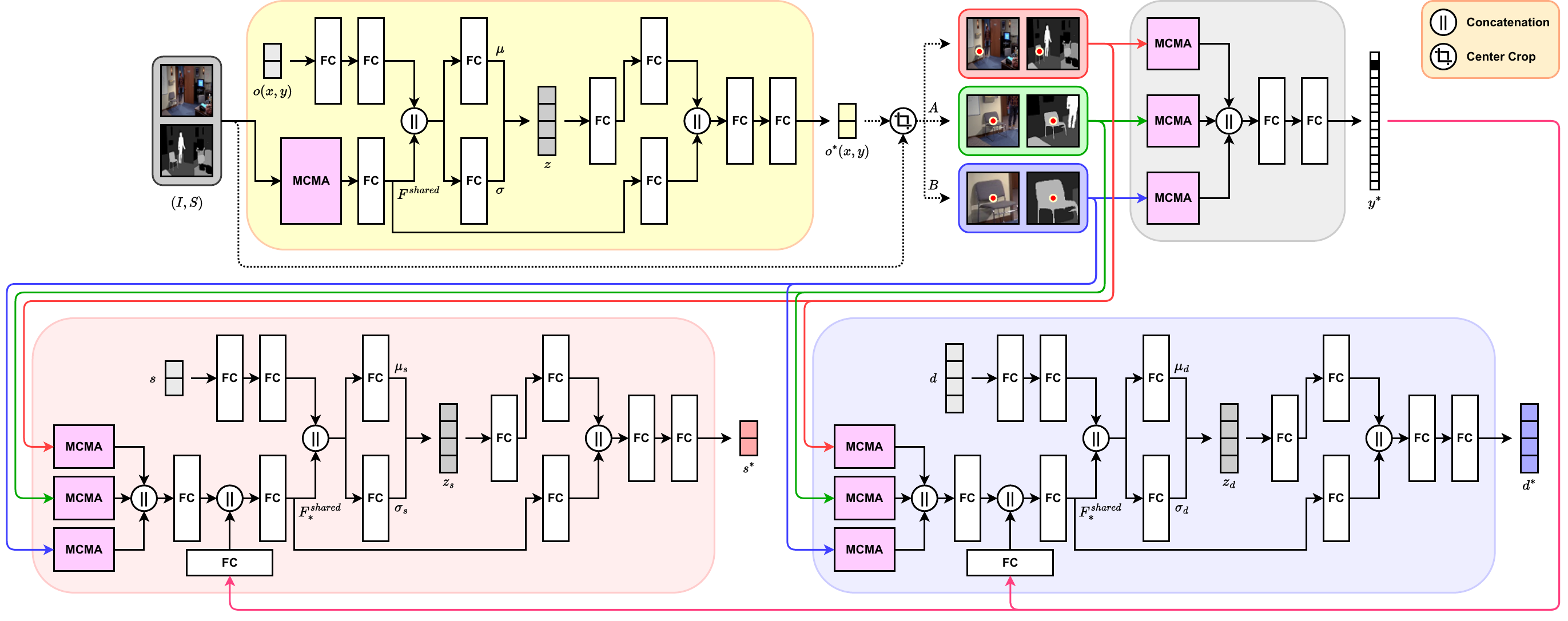}
  \caption{An illustration of the proposed architecture. The workflow is divided into four subnetworks to estimate the probable location $o^*$, pose template class $y^*$, scaling parameters $s^*$, and linear deformations $d^*$ of a potential target pose. Every subnetwork exclusively uses the proposed \emph{Mutual Cross-Modal Attention} (\textbf{MCMA}) block to encode global and local scene contexts as shown in Fig. \ref{fig:mcma_block}.}
  \label{fig:architecture}
\end{figure*}

\section{Experiments}
\label{sec:experiments}

\begin{figure*}[t]
  \centering
  \includegraphics[width=\linewidth]{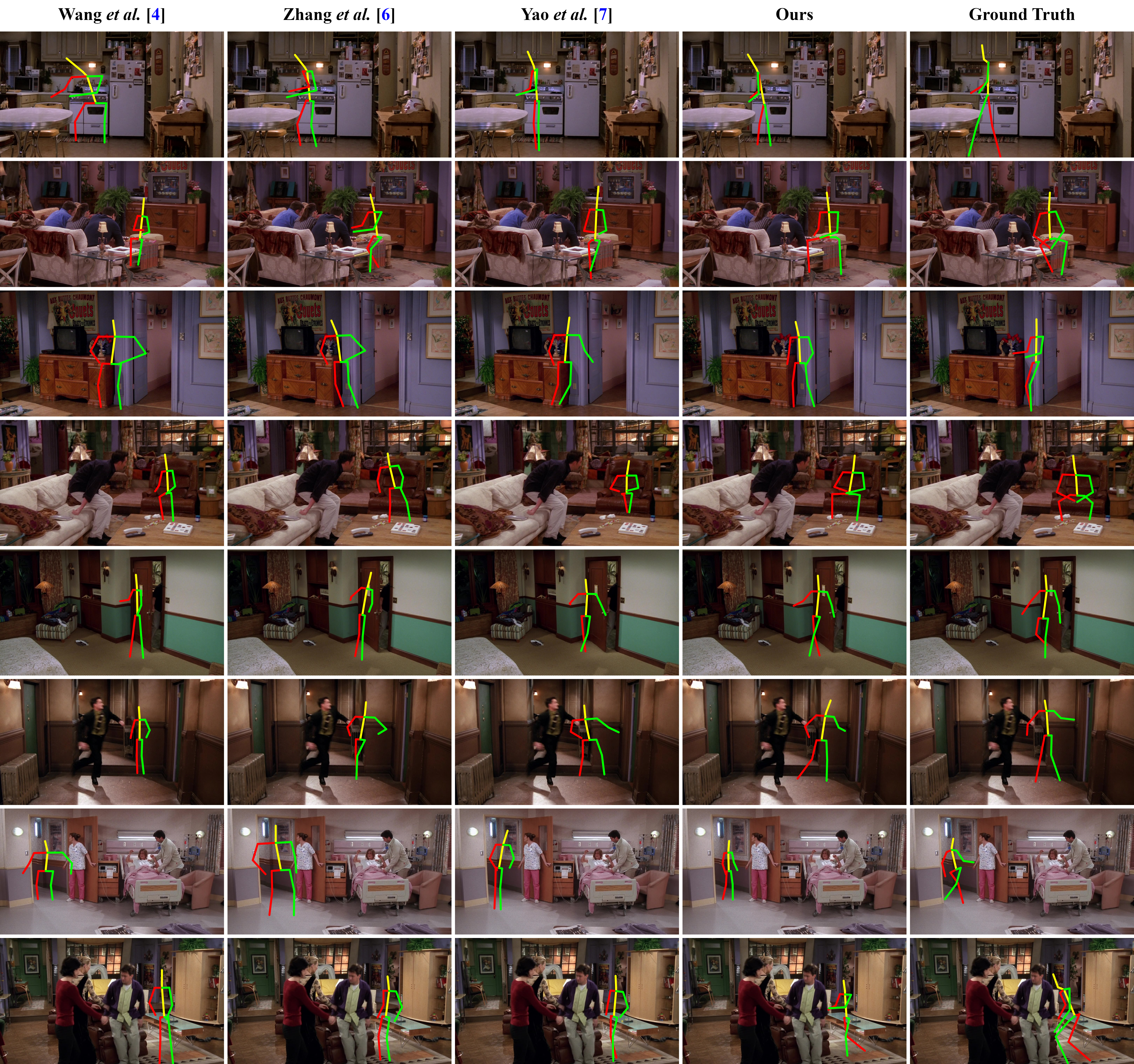}
  \caption{Qualitative comparison of the proposed method with existing human affordance generation techniques by Wang \emph{et al.} \cite{wang2017binge}, Zhang \emph{et al.} \cite{zhang2022inpaint}, and Yao \emph{et al.} \cite{yao2023scene}.}
  \label{fig:comparison}
  \vspace{-1.0em}
\end{figure*}

\subsection{Dataset}
\label{sec:experiments_dataset}
The number of large-scale annotated public datasets for complex 2D affordance generation is significantly limited in the literature. The main challenge is to obtain a specific frame in two different states -- \emph{with} and \emph{without} populated with random persons. Researchers \cite{zhang2022inpaint, zhu2023topnet} have attempted to resolve the issue by randomly removing and inpainting existing person instances from the scene. However, these datasets \cite{zhang2022inpaint, zhu2023topnet} are not publicly available, and our attempt to remove persons from a complex scene leaves significant visual artifacts even with state-of-the-art inpainting techniques \cite{suvorov2022resolution}, making such data generation methods \cite{zhang2022inpaint, zhu2023topnet} insufficient for our purpose.

Following the recent works \cite{wang2017binge, yao2023scene}, we train and evaluate the proposed method on an openly available large-scale sitcom dataset \cite{wang2017binge} for fair comparisons. The dataset comprises 28837 human interaction samples over 11499 video frames extracted from seven sitcoms. The training set consists of 24967 poses over 10009 frames from six sitcoms, while the evaluation set contains 3870 poses over 1490 frames from one sitcom. Each human pose is represented with 16 keypoints following the standard MPII format \cite{andriluka142d}.

\subsection{Visual results and evaluation metrics}
\label{sec:experiments_results}
\noindent
\textbf{Qualitative analysis:} To demonstrate the efficacy of the proposed method, we compare the visual results against major existing works \cite{wang2017binge, yao2023scene, zhang2022inpaint} on 2D human affordance generation. In \cite{wang2017binge, yao2023scene}, the networks directly learn from a single modality of image features. In \cite{zhang2022inpaint}, the authors introduce an adversarial learning mechanism with two additional modalities of segmentation and depth maps alongside image features. We notice that adopting a transformer-based architecture \cite{yao2023scene} or combining multiple modalities in an adversarial learning method \cite{zhang2022inpaint} provide only marginal improvements over the baseline approach \cite{wang2017binge}. In contrast, the proposed method focuses on imposing a better semantic constraint in the learning strategy by introducing a novel cross-attention mechanism. Fig. \ref{fig:comparison} illustrates a qualitative comparison of the proposed method with existing human affordance generation techniques \cite{wang2017binge, yao2023scene, zhang2022inpaint}. The visual results show that our approach produces more realistic human interactions in complex scenes than previous methods.

\noindent
\textbf{Quantitative evaluation:}
We evaluate the alignment of the generated pose with the ground truth for analytically comparing the proposed method against existing human affordance generation techniques \cite{wang2017binge, yao2023scene, zhang2022inpaint}. In particular, we use the percentage of correct keypoints (PCK/PCKh) \cite{yang2012articulated}, average keypoint distance (AKD), mean absolute error (MAE), mean squared error (MSE), and cosine similarity (SIM) as the evaluation metrics for pose alignment. To estimate the correctness of the estimated scale, we measure the intersection over union (IOU) between the target and predicted pose bounding boxes. Following \cite{yao2023scene}, our comparative study also includes evaluation results from additional baseline methods (heatmap and regression), pose estimation techniques \cite{artacho2020unipose, li2021pose}, and object placement algorithms \cite{zhang2020learning, zhou2022learning}. Additionally, we train and evaluate another variation of the proposed architecture by replacing \emph{\textbf{semantic segmentation maps}} with \emph{\textbf{depth maps}}. \textbf{PCK} and \textbf{PCKh} measure the similarity between two poses by computing the fraction of correctly aligned keypoints, where a valid alignment denotes that the distance between two respective keypoints is within a predetermined tolerance. PCK uses $\alpha * torso~width$, while PCKh uses $\beta * head~size$ as the tolerance threshold, where $0 < \alpha, \beta \leqslant 1$. \textbf{AKD} is the average Euclidean distance between respective pairs of target and predicted keypoints. \textbf{MAE} and \textbf{MSE} measure the average absolute and squared linear deviations, respectively, along both coordinate axes between all respective pairs of actual and inferred keypoints. \textbf{SIM} evaluates the average cosine similarity between the positional vectors corresponding to every target and predicted keypoint pair. \textbf{IOU} enumerates the intersection over union ratio between the bounding rectangles of predicted and target poses. We summarize the quantitative evaluation scores for each competing method in Table \ref{tab:comparison}. The proposed method exhibits significantly better evaluation scores, reflecting the apparent visual superiority of the qualitative analysis.

\noindent
\textbf{Subjective evaluation (User study):}
While the evaluation metrics are analogous to visual rationality for most cases in our experiments, the quantitative scores alone may not be sufficient to claim the superiority of a generation scheme. The reason is the potential uncertainty of a generated human pose. Within the given scene context, it is possible to obtain multiple valid human interactions with the environment. Therefore, the evaluation scores against a specific ground truth pose may not always provide a rational judgment of superiority. We have conducted an opinion-based user study with 42 volunteers to select the most visually realistic sample from a pool of images generated by the competing methods, including ground truth. The mean opinion score \textbf{(MOS)} is estimated as the average fraction of times a method is preferred over the others. As shown in Table \ref{tab:comparison}, the proposed approach achieves a significantly higher user preference over other existing methods, indicating the best semantic integrity in the generated samples similar to the ground truth.

\begin{table*}[t]
\centering
\caption{Quantitative comparison of the proposed method with existing pose estimation \cite{artacho2020unipose, li2021pose}, object placement \cite{zhang2020learning, zhou2022learning}, and affordance generation \cite{wang2017binge, yao2023scene, zhang2022inpaint} techniques. The best scores are in \textbf{bold}, and the second-best scores are \underline{underlined}.}
\label{tab:comparison}
\begin{tabular}{lcccccccc}
\hline
\rowcolor[HTML]{ECECEC}
  \textbf{Method} &
  \textbf{PCK $\uparrow$} &
  \textbf{PCKh $\uparrow$} &
  \textbf{AKD $\downarrow$} &
  \textbf{MAE $\downarrow$} &
  \textbf{MSE $\downarrow$} &
  \textbf{SIM $\uparrow$} &
  \textbf{IOU $\uparrow$} &
  \textbf{\textcolor{blue}{User Score $\uparrow$}} \\ \hline
Heatmap                                     & 0.363 & 0.422 & 11.298 & ~7.112 & ~53.45 & 0.9864 & 0.402 & 0.000 \\
Regression                                  & 0.386 & 0.451 & 10.929 & ~6.840 & ~51.29 & 0.9899 & 0.426 & 0.000 \\
UniPose \cite{artacho2020unipose}           & 0.387 & 0.447 & ~9.966 & ~6.241 & ~46.78 & 0.9918 & 0.471 & 0.012 \\
PRTR \cite{li2021pose}                      & 0.408 & 0.474 & ~9.724 & ~6.088 & ~45.72 & 0.9934 & 0.489 & 0.025 \\
PlaceNet \cite{zhang2020learning}           & 0.060 & 0.072 & 79.978 & 50.325 & 377.78 & 0.9476 & 0.118 & 0.000 \\
GracoNet \cite{zhou2022learning}            & 0.380 & 0.441 & 10.576 & ~6.614 & ~49.60 & 0.9912 & 0.427 & 0.000 \\ \hline
Wang \emph{et al.} \cite{wang2017binge}     & 0.401 & 0.462 & ~9.940 & ~6.208 & ~46.65 & 0.9928 & 0.482 & 0.022 \\
Zhang \emph{et al.} \cite{zhang2022inpaint} & 0.372 & 0.428 & 10.252 & ~6.409 & ~48.14 & 0.9906 & 0.405 & 0.005 \\
Yao \emph{et al.} \cite{yao2023scene}       &
\underline{0.414} &
\underline{0.479} &
~9.514 &
~5.918 &
~44.86 &
0.9954 &
0.494 &
0.104 \\ \hline
\rowcolor[HTML]{FFFFCC}
\textbf{Ours \emph{(Depth)}} &
0.407 &
0.472 &
\underline{~6.680} &
\underline{~4.163} &
\underline{~32.78} &
\underline{0.9966} &
\underline{0.566} &
\underline{0.205} \\
\rowcolor[HTML]{FFFFCC}
\textbf{Ours \emph{(Semantic)}} &
\textbf{0.433} &
\textbf{0.503} &
\textbf{~6.352} &
\textbf{~3.972} &
\textbf{~29.81} &
\textbf{0.9969} &
\textbf{0.566} &
\textbf{0.299} \\ \hline \hline
Ground Truth                                & 1.000 & 1.000 & ~0.000 & ~0.000 & ~~0.00 & 1.0000 & 1.000 & 0.328 \\ \hline
\end{tabular}%
\end{table*}

\begin{table*}[t]
\centering
\caption{Quantitative ablation analysis of different variations of the proposed network architecture.}
\label{tab:ablation}
\resizebox{\textwidth}{!}{%
\begin{tabular}{llccccccc}
\hline
\rowcolor[HTML]{ECECEC}
  \textbf{Model} &
  \textbf{Context} &
  \textbf{PCK $\uparrow$} &
  \textbf{PCKh $\uparrow$} &
  \textbf{AKD $\downarrow$} &
  \textbf{MAE $\downarrow$} &
  \textbf{MSE $\downarrow$} &
  \textbf{SIM $\uparrow$} &
  \textbf{IOU $\uparrow$} \\ \hline
\texttt{\textbf{Baseline}} & None &
0.274 \textcolor{gray}{$\pm$0.006} & 0.322 \textcolor{gray}{$\pm$0.008} & 9.998 \textcolor{gray}{$\pm$0.042} & 6.248 \textcolor{gray}{$\pm$0.034} & 48.920 \textcolor{gray}{$\pm$0.408} & 0.9845 \textcolor{gray}{$\pm$0.0005} & 0.398 \textcolor{gray}{$\pm$0.004} \\ \hline
\texttt{\textbf{Self-I}} & Image &
0.346 \textcolor{gray}{$\pm$0.004} & 0.399 \textcolor{gray}{$\pm$0.005} & 7.899 \textcolor{gray}{$\pm$0.035} & 4.925 \textcolor{gray}{$\pm$0.025} & 38.714 \textcolor{gray}{$\pm$0.406} & 0.9947 \textcolor{gray}{$\pm$0.0004} & 0.458 \textcolor{gray}{$\pm$0.012} \\ \hline
\texttt{\textbf{Self-D}} & Depth &
0.317 \textcolor{gray}{$\pm$0.004} & 0.372 \textcolor{gray}{$\pm$0.006} & 8.582 \textcolor{gray}{$\pm$0.025} & 5.625 \textcolor{gray}{$\pm$0.028} & 42.104 \textcolor{gray}{$\pm$0.344} & 0.9888 \textcolor{gray}{$\pm$0.0002} & 0.406 \textcolor{gray}{$\pm$0.002} \\
\texttt{\textbf{Cross-D/I}} & Image &
0.341 \textcolor{gray}{$\pm$0.004} & 0.387 \textcolor{gray}{$\pm$0.008} & 8.146 \textcolor{gray}{$\pm$0.032} & 5.077 \textcolor{gray}{$\pm$0.028} & 39.972 \textcolor{gray}{$\pm$0.228} & 0.9927 \textcolor{gray}{$\pm$0.0005} & 0.447 \textcolor{gray}{$\pm$0.006} \\
\texttt{\textbf{Cross-I/D}} & Depth &
0.374 \textcolor{gray}{$\pm$0.005} & 0.429 \textcolor{gray}{$\pm$0.005} & 7.260 \textcolor{gray}{$\pm$0.047} & 4.424 \textcolor{gray}{$\pm$0.025} & 35.649 \textcolor{gray}{$\pm$0.301} & 0.9952 \textcolor{gray}{$\pm$0.0002} & 0.498 \textcolor{gray}{$\pm$0.002} \\
\rowcolor[HTML]{FFCCCC}
\texttt{\textbf{Mutual-I+D}} & Image + Depth &
0.407 \textcolor{gray}{$\pm$0.003} & 0.472 \textcolor{gray}{$\pm$0.002} & 6.680 \textcolor{gray}{$\pm$0.031} & 4.163 \textcolor{gray}{$\pm$0.020} & 32.782 \textcolor{gray}{$\pm$0.294} & 0.9966 \textcolor{gray}{$\pm$0.0001} & 0.566 \textcolor{gray}{$\pm$0.004} \\ \hline
\texttt{\textbf{Self-S}} & Semantic &
0.336 \textcolor{gray}{$\pm$0.006} & 0.391 \textcolor{gray}{$\pm$0.009} & 8.174 \textcolor{gray}{$\pm$0.038} & 5.112 \textcolor{gray}{$\pm$0.024} & 38.452 \textcolor{gray}{$\pm$0.404} & 0.9895 \textcolor{gray}{$\pm$0.0001} & 0.418 \textcolor{gray}{$\pm$0.014} \\
\texttt{\textbf{Cross-S/I}} & Image &
0.355 \textcolor{gray}{$\pm$0.005} & 0.407 \textcolor{gray}{$\pm$0.008} & 7.746 \textcolor{gray}{$\pm$0.039} & 4.904 \textcolor{gray}{$\pm$0.018} & 36.354 \textcolor{gray}{$\pm$0.477} & 0.9950 \textcolor{gray}{$\pm$0.0002} & 0.478 \textcolor{gray}{$\pm$0.008} \\
\texttt{\textbf{Cross-I/S}} & Semantic &
0.398 \textcolor{gray}{$\pm$0.006} & 0.458 \textcolor{gray}{$\pm$0.004} & 6.904 \textcolor{gray}{$\pm$0.045} & 4.325 \textcolor{gray}{$\pm$0.029} & 32.402 \textcolor{gray}{$\pm$0.344} & 0.9958 \textcolor{gray}{$\pm$0.0002} & 0.515 \textcolor{gray}{$\pm$0.005} \\
\rowcolor[HTML]{CCFFCC}
\texttt{\textbf{Mutual-I+S}} & Image + Semantic &
\textbf{0.433} \textcolor{gray}{$\pm$0.004} & \textbf{0.503} \textcolor{gray}{$\pm$0.004} & \textbf{6.352} \textcolor{gray}{$\pm$0.036} & \textbf{3.972} \textcolor{gray}{$\pm$0.022} & \textbf{29.810} \textcolor{gray}{$\pm$0.326} & \textbf{0.9969} \textcolor{gray}{$\pm$0.0001} & \textbf{0.566} \textcolor{gray}{$\pm$0.002} \\ \hline 
\end{tabular}%
}
\end{table*}

\begin{figure}[t]
\centering
\captionsetup[subfloat]{labelfont=bf}
\subfloat{\includegraphics[width=0.32\linewidth]{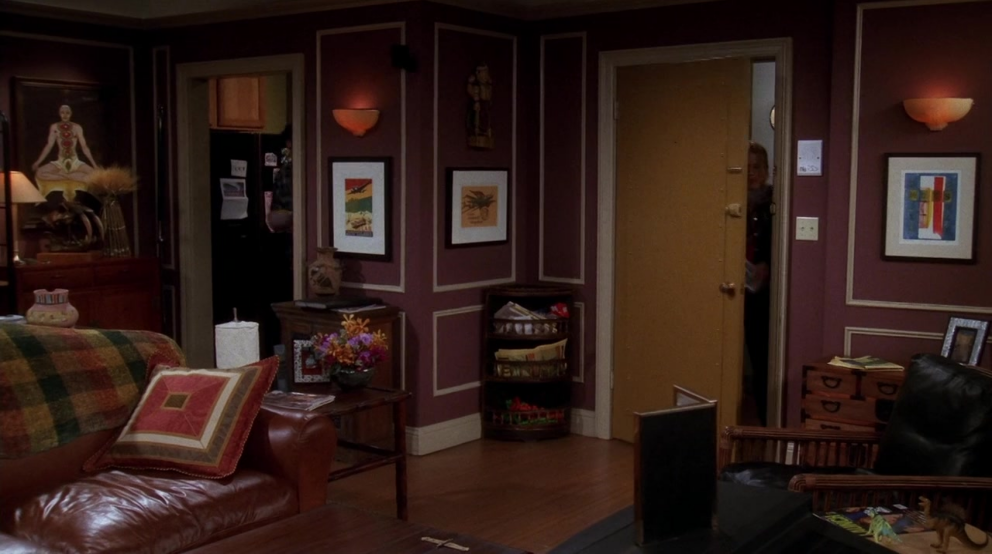}}\hfil
\subfloat{\includegraphics[width=0.32\linewidth]{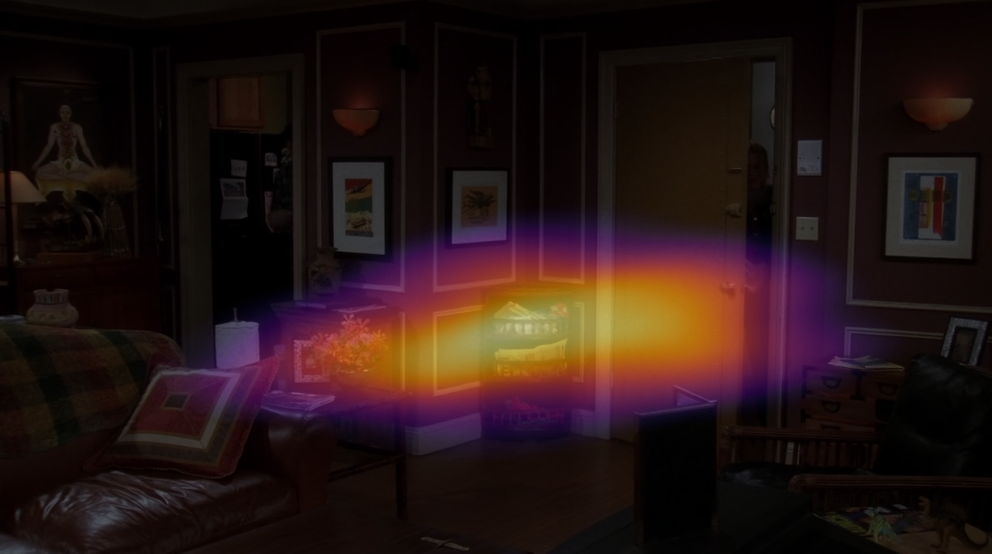}}\hfil
\subfloat{\includegraphics[width=0.32\linewidth]{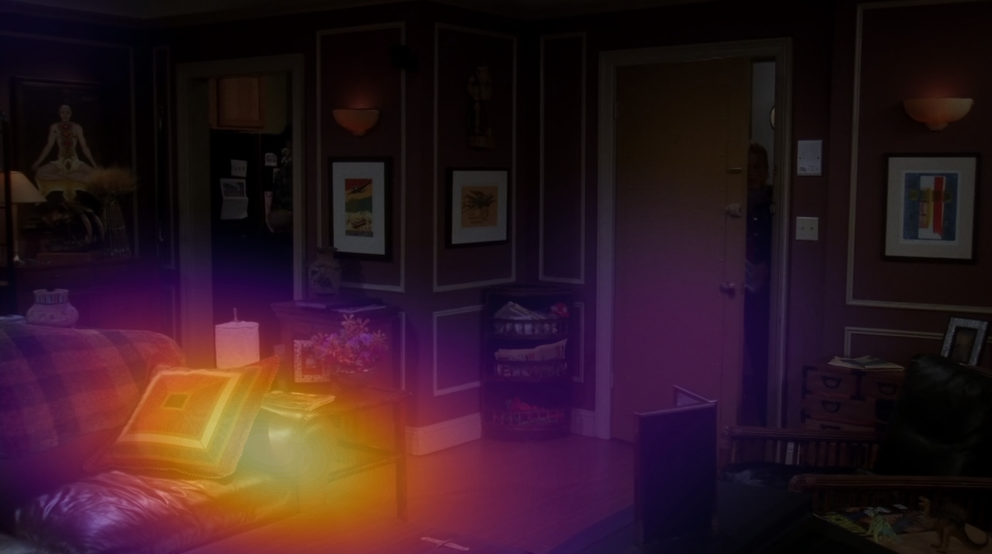}}
\vspace{-0.75em}
\subfloat{\includegraphics[width=0.32\linewidth]{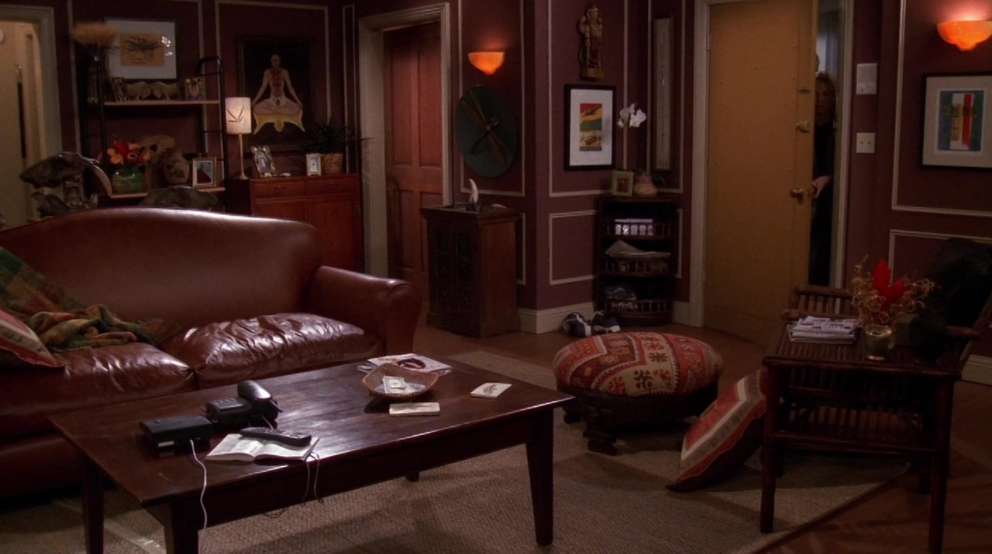}}\hfil
\subfloat{\includegraphics[width=0.32\linewidth]{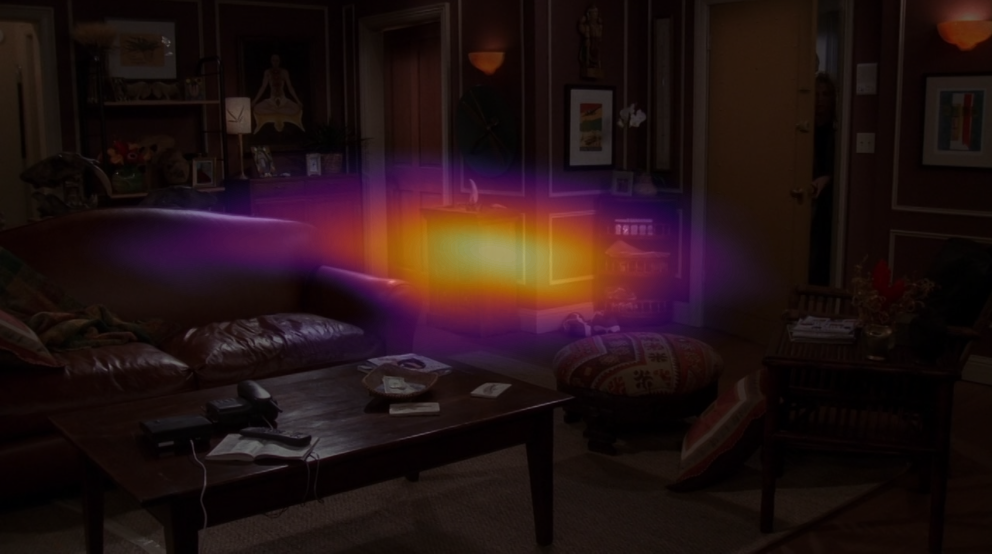}}\hfil
\subfloat{\includegraphics[width=0.32\linewidth]{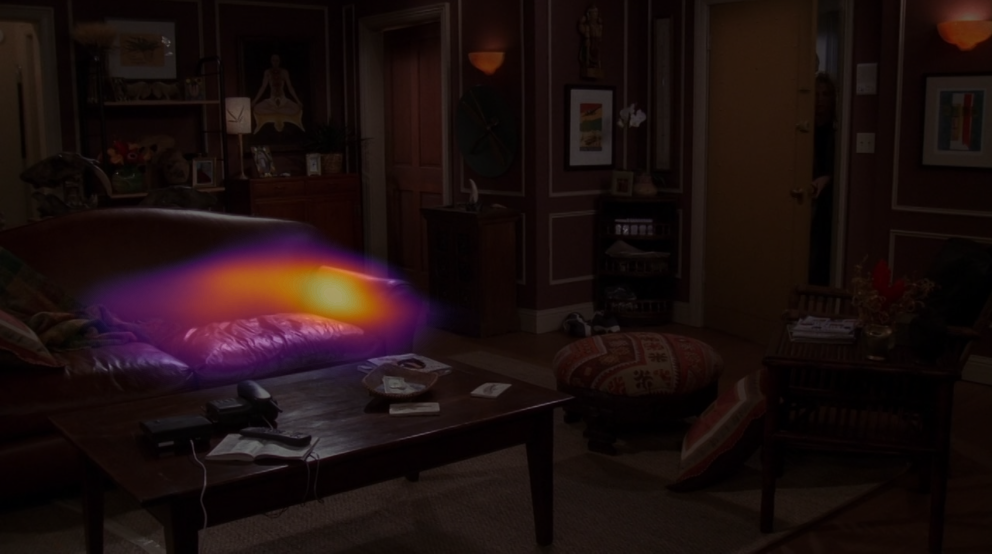}}
\vspace{-0.75em}
\subfloat{\includegraphics[width=0.32\linewidth]{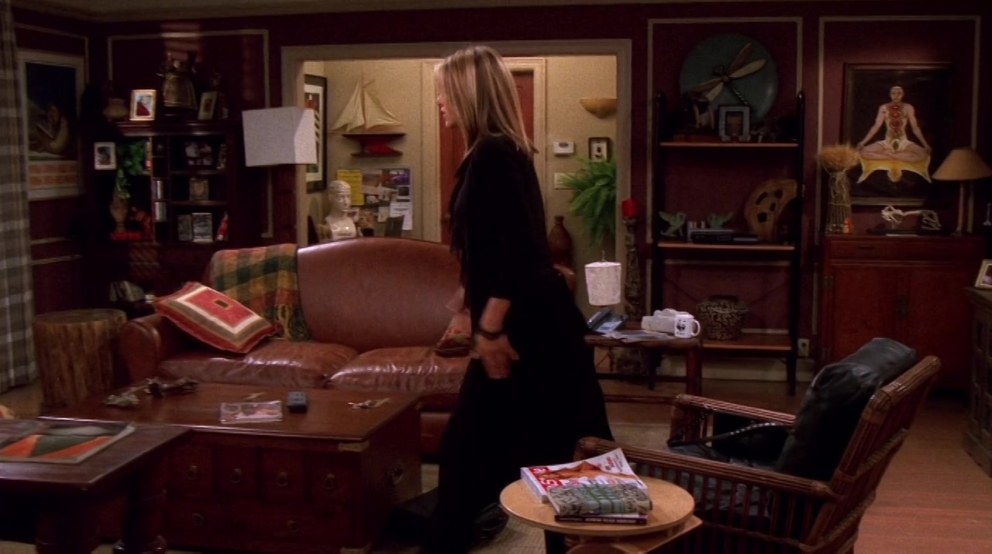}}\hfil
\subfloat{\includegraphics[width=0.32\linewidth]{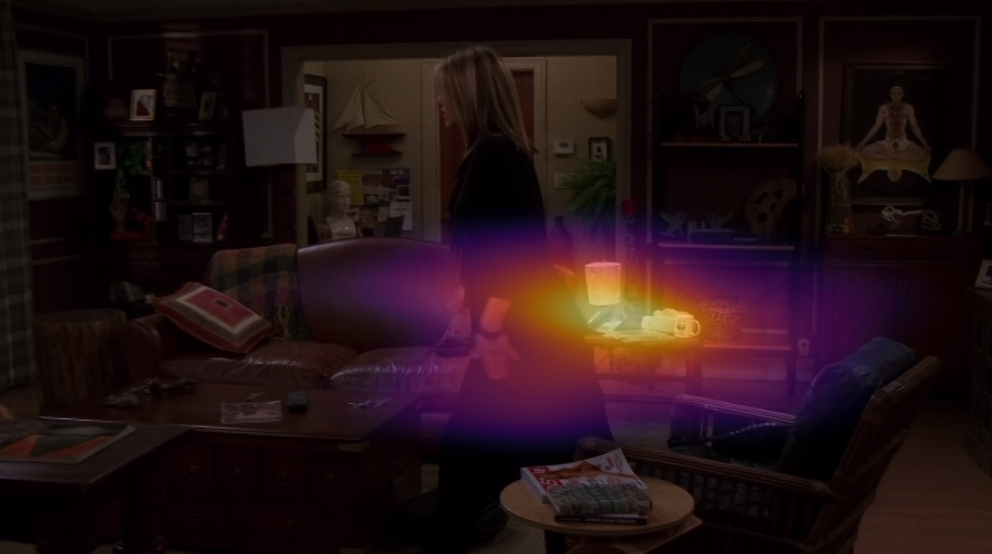}}\hfil
\subfloat{\includegraphics[width=0.32\linewidth]{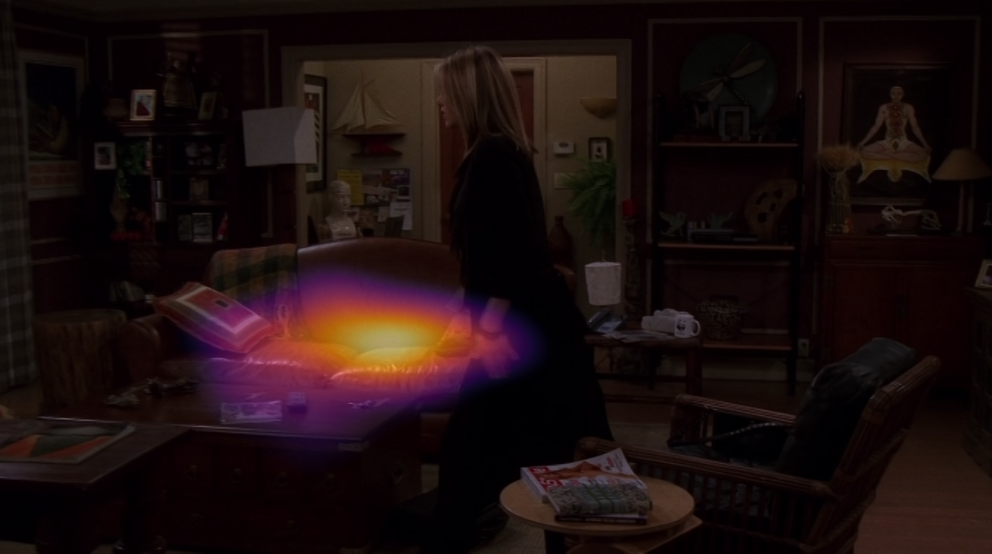}}
\caption{\textcolor{black}{Visualization of the learned distribution. \textbf{(Left)} Input scene. \textbf{(Middle)} Distribution of \emph{standing} poses. \textbf{(Right)} Distribution of \emph{sitting} poses.}}
\label{fig:distribution}
\vspace{-1.0em}
\end{figure}

\noindent
\textcolor{black}{\textbf{Visualization of the learned distribution:}
Due to many feasible outcomes, there are substantial ambiguities when inferring a possible pose at a random location within the scene, depending on the surrounding context. The ground truth is only one of the many possibilities. For example, a selected position in front of a chair can lead to either \emph{standing} or \emph{sitting} postures. Therefore, the association between an estimated pose and scene objects is an ideal way to visualize the sampled pose distribution. We discover two broad pose categories, \emph{standing} and \emph{sitting}, in the dataset by manually inspecting the given templates. To visualize the learned distribution, we randomly sample 10000 poses for a scene and assign a pose category (\emph{standing} or \emph{sitting}) to each sampled location based on the predicted pose template at that position. Then, we visualize the bivariate distribution of the sampled coordinates for each pose category. Fig. \ref{fig:distribution} shows a few such visualizations, exhibiting ideal associations between a pose type and the scene objects.}

\subsection{Ablation study}
\label{sec:experiments_ablation}

\begin{figure*}[t]
\centering
\captionsetup[subfloat]{labelfont=bf}
\subfloat[]{\includegraphics[width=0.16\linewidth]{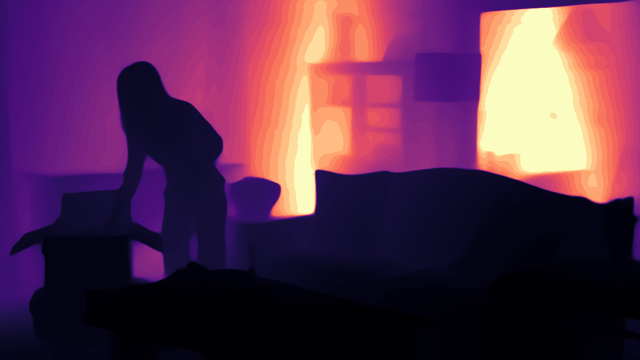}}\hfil
\subfloat[]{\includegraphics[width=0.16\linewidth]{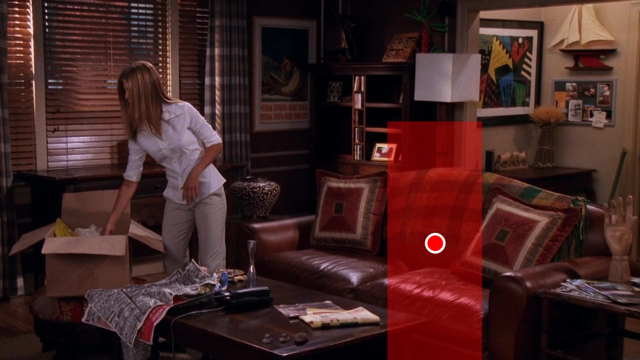}}\hfil
\subfloat[]{\includegraphics[width=0.16\linewidth]{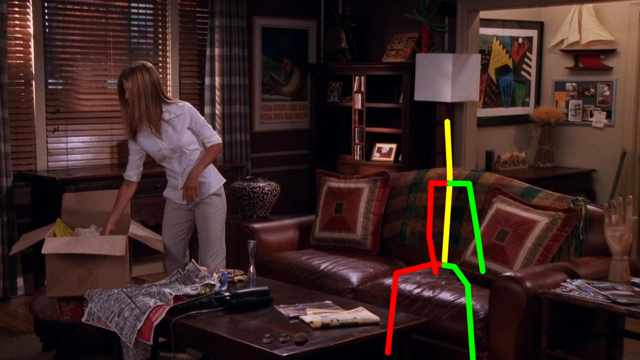}}\hfil
\subfloat[]{\includegraphics[width=0.16\linewidth]{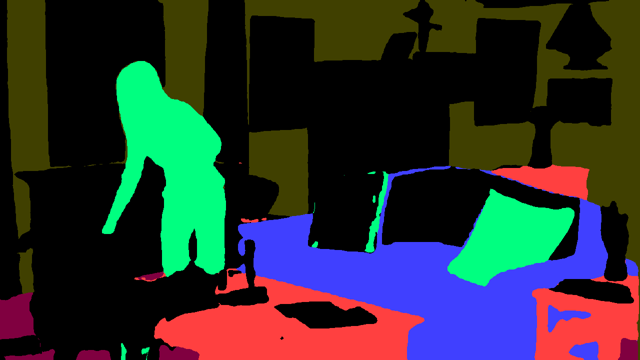}}\hfil
\subfloat[]{\includegraphics[width=0.16\linewidth]{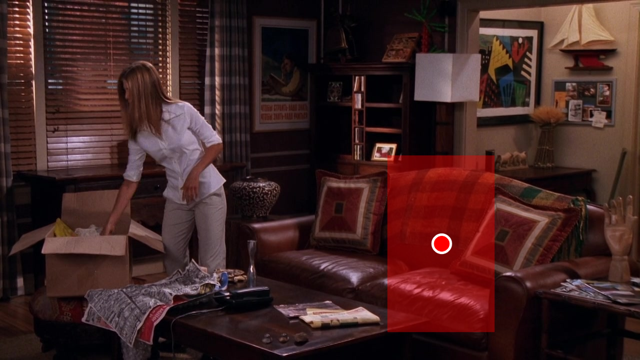}}\hfil
\subfloat[]{\includegraphics[width=0.16\linewidth]{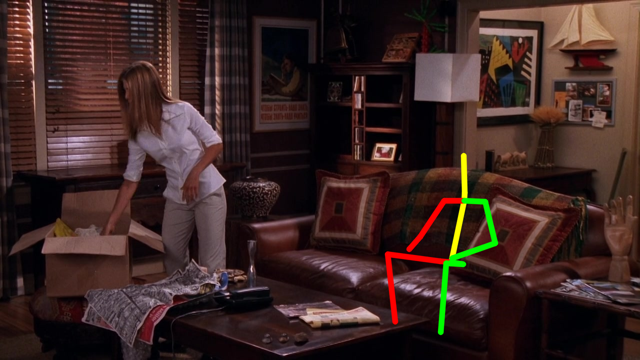}}
\vspace{-0.75em}
\subfloat{\includegraphics[width=0.16\linewidth]{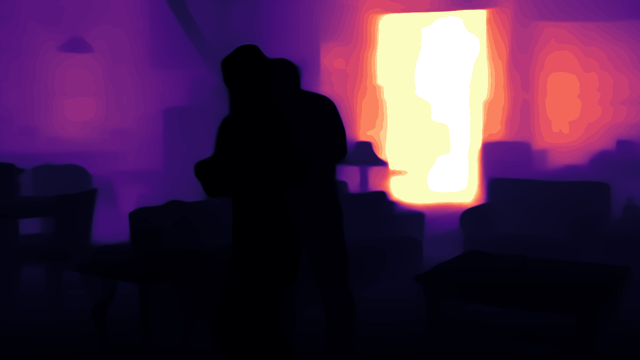}}\hfil
\subfloat{\includegraphics[width=0.16\linewidth]{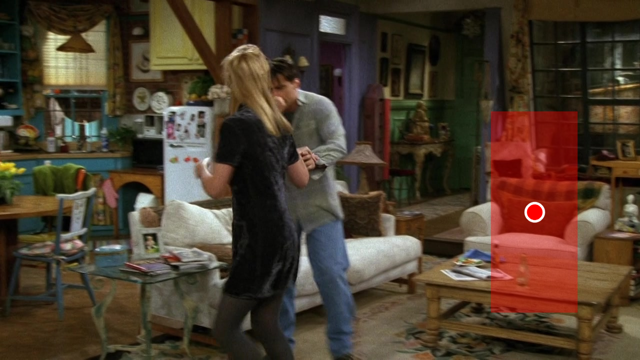}}\hfil
\subfloat{\includegraphics[width=0.16\linewidth]{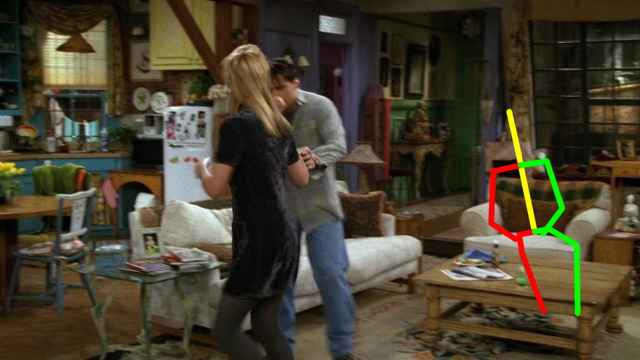}}\hfil
\subfloat{\includegraphics[width=0.16\linewidth]{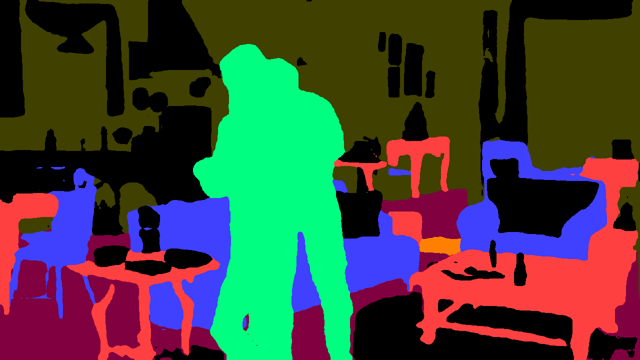}}\hfil
\subfloat{\includegraphics[width=0.16\linewidth]{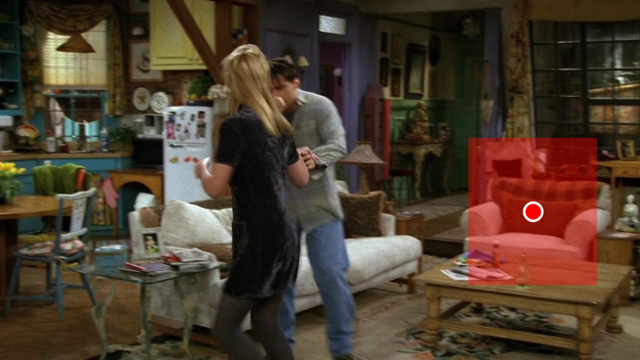}}\hfil
\subfloat{\includegraphics[width=0.16\linewidth]{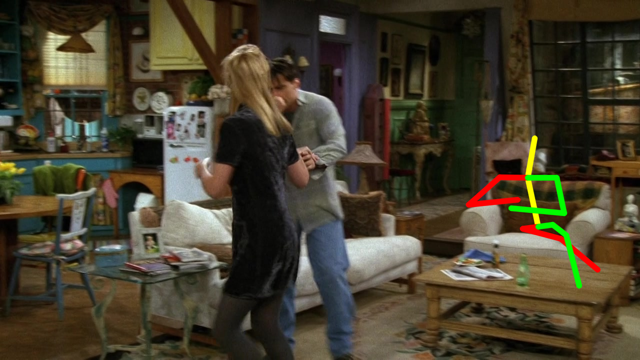}}
\vspace{-0.75em}
\subfloat{\includegraphics[width=0.16\linewidth]{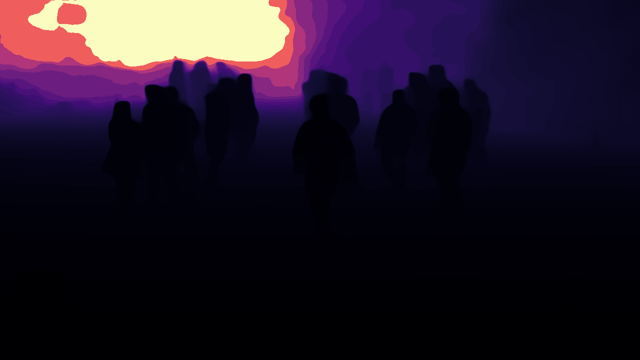}}\hfil
\subfloat{\includegraphics[width=0.16\linewidth]{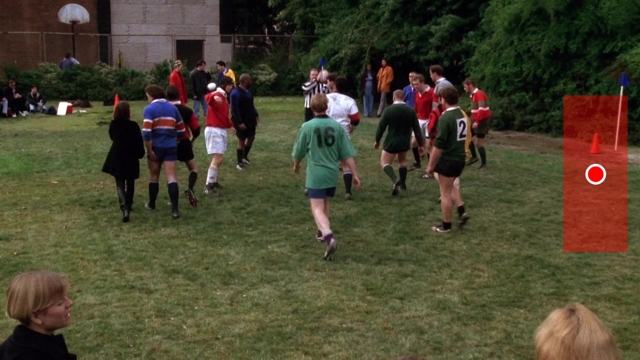}}\hfil
\subfloat{\includegraphics[width=0.16\linewidth]{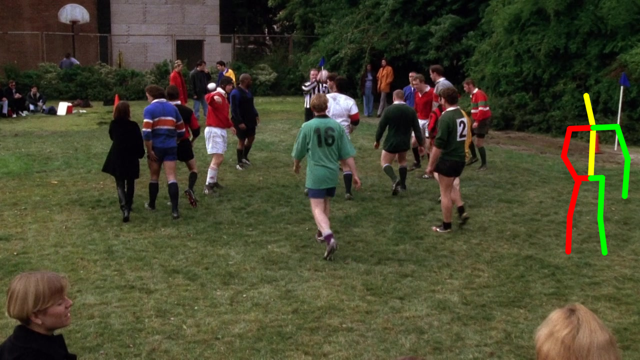}}\hfil
\subfloat{\includegraphics[width=0.16\linewidth]{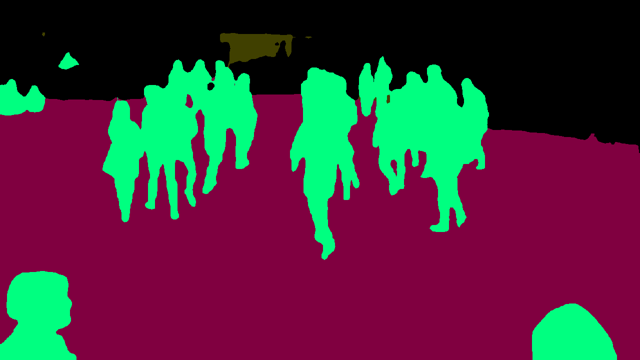}}\hfil
\subfloat{\includegraphics[width=0.16\linewidth]{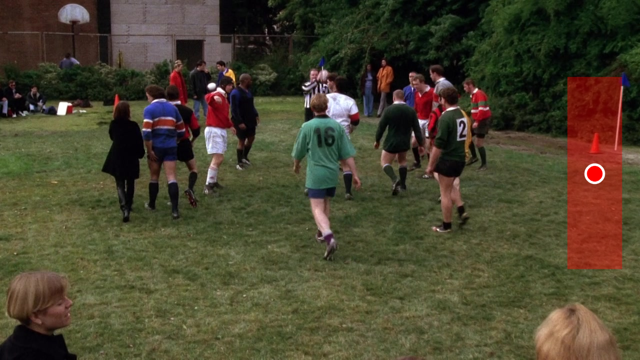}}\hfil
\subfloat{\includegraphics[width=0.16\linewidth]{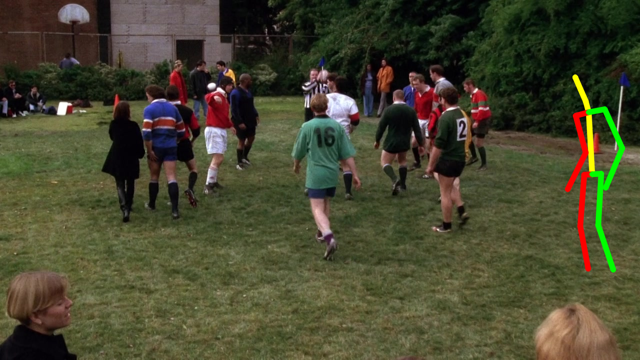}}
\vspace{-0.75em}
\subfloat{\includegraphics[width=0.16\linewidth]{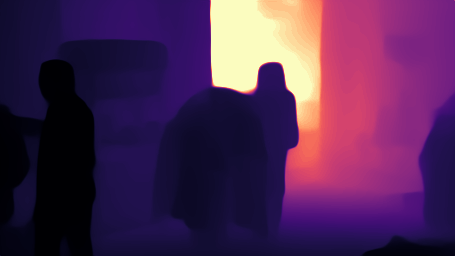}}\hfil
\subfloat{\includegraphics[width=0.16\linewidth]{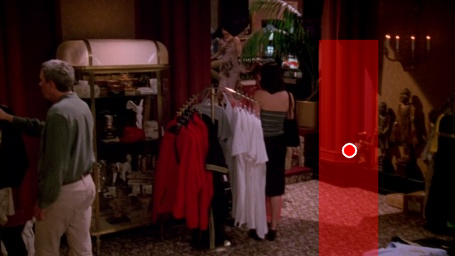}}\hfil
\subfloat{\includegraphics[width=0.16\linewidth]{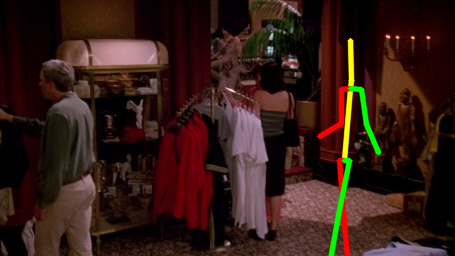}}\hfil
\subfloat{\includegraphics[width=0.16\linewidth]{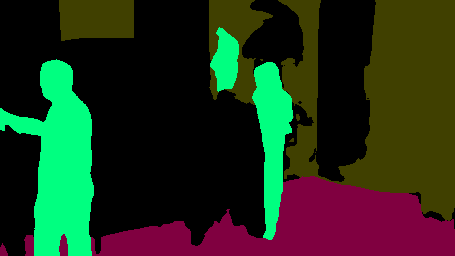}}\hfil
\subfloat{\includegraphics[width=0.16\linewidth]{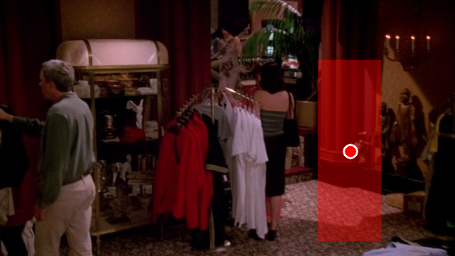}}\hfil
\subfloat{\includegraphics[width=0.16\linewidth]{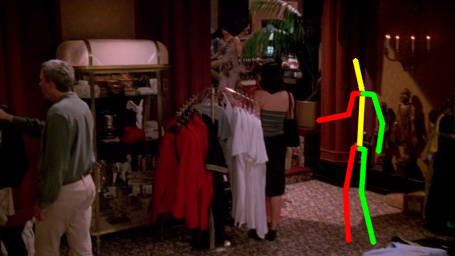}}
\vspace{-0.75em}
\subfloat{\includegraphics[width=0.16\linewidth]{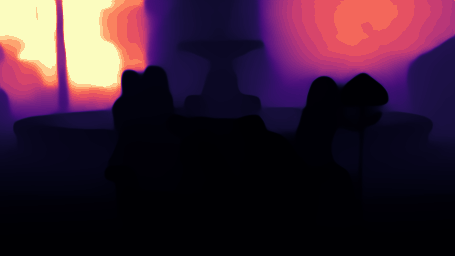}}\hfil
\subfloat{\includegraphics[width=0.16\linewidth]{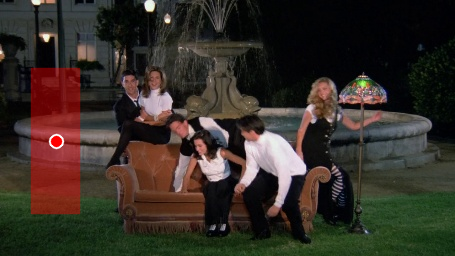}}\hfil
\subfloat{\includegraphics[width=0.16\linewidth]{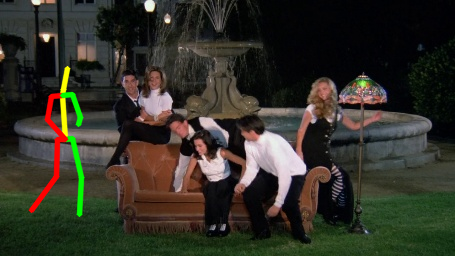}}\hfil
\subfloat{\includegraphics[width=0.16\linewidth]{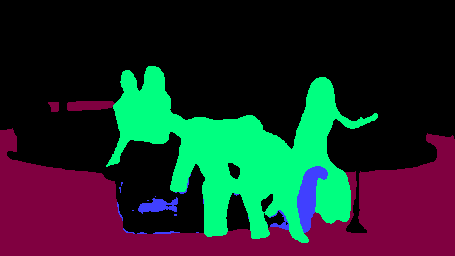}}\hfil
\subfloat{\includegraphics[width=0.16\linewidth]{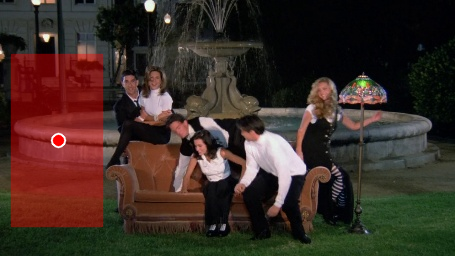}}\hfil
\subfloat{\includegraphics[width=0.16\linewidth]{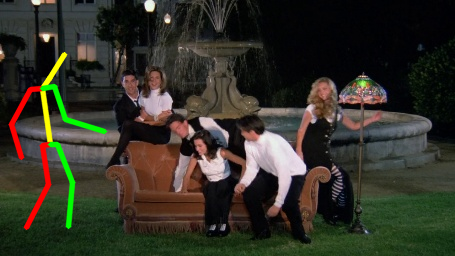}}
\vspace{-0.75em}
\subfloat{\includegraphics[width=0.16\linewidth]{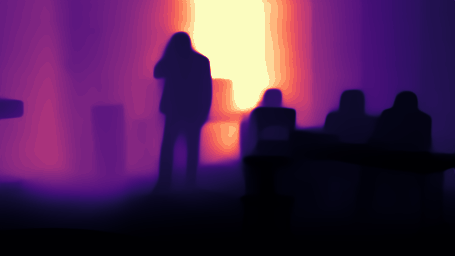}}\hfil
\subfloat{\includegraphics[width=0.16\linewidth]{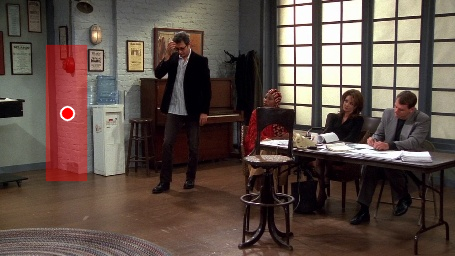}}\hfil
\subfloat{\includegraphics[width=0.16\linewidth]{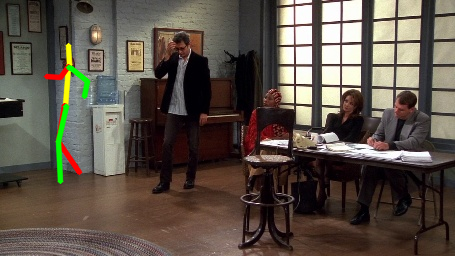}}\hfil
\subfloat{\includegraphics[width=0.16\linewidth]{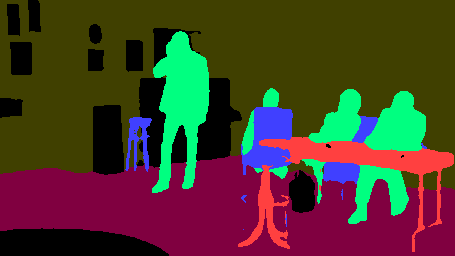}}\hfil
\subfloat{\includegraphics[width=0.16\linewidth]{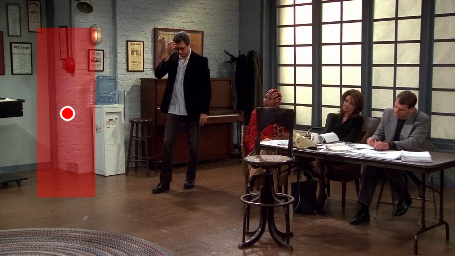}}\hfil
\subfloat{\includegraphics[width=0.16\linewidth]{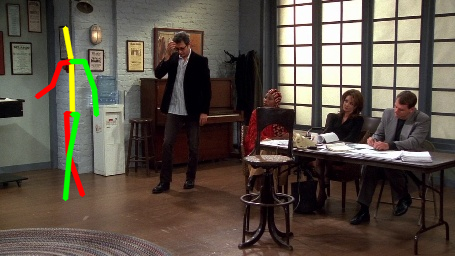}}
\caption{Qualitative ablation analysis of the proposed network architecture with different input modalities. \textbf{(a)} Depth map. \textbf{(b)} Estimated bounding region from depth context. \textbf{(c)} Estimated pose from depth context. \textbf{(d)} Semantic map. \textbf{(e)} Estimated bounding region from semantic context. \textbf{(f)} Estimated pose from semantic context.}
\label{fig:ablation}
\end{figure*}

In the ablation analysis, we evaluate ten different network configurations by varying attention mechanisms in the proposed MCMA block and altering contextual modalities (\emph{depth map} or \emph{semantic map}) to determine an optimal network architecture. The \textbf{\texttt{Baseline}} architecture entirely excludes contextual supervision by removing all MCMA blocks from the proposed architecture. Configurations \textbf{\texttt{Self-I}}, \textbf{\texttt{Self-D}}, and \textbf{\texttt{Self-S}} also drop all MCMA blocks but introduce \emph{self-attention} on a single modality. In particular, \textbf{\texttt{Self-I}} uses self-attention on the image $I$ itself, while \textbf{\texttt{Self-D}} and \textbf{\texttt{Self-S}} use self-attention on the \emph{depth map} $D$ and \emph{semantic segmentation map} $S$, respectively. The next four architectures \textbf{\texttt{Cross-D/I}}, \textbf{\texttt{Cross-I/D}}, \textbf{\texttt{Cross-S/I}}, and \textbf{\texttt{Cross-I/S}} introduce \emph{cross-attention} from an additional input stream. Specifically, for configurations \textbf{\texttt{Cross-D/I}} and \textbf{\texttt{Cross-S/I}}, we compute query ($Q$) from $D$ and $S$, respectively, while retrieving key ($K$) and value ($V$) from $I$. Likewise, for configurations \textbf{\texttt{Cross-I/D}} and \textbf{\texttt{Cross-I/S}}, we compute $Q$ from $I$, while estimating $K$ and $V$ from the contextual input $D$ or $S$, respectively. The final two network variants \textbf{\texttt{Mutual-D}} and \textbf{\texttt{Mutual-S}} use the proposed \emph{mutual cross-modal attention} (MCMA) mechanism utilizing $D$ or $S$ as the auxiliary contextual supervision, respectively, alongside $I$. In all our experiments, we estimate the \emph{depth maps} using a recent monocular depth estimation technique \emph{Depth Anything} \cite{yang2024depth} and the \emph{semantic segmentation maps} using a recent transformer-based universal image segmentation method \emph{OneFormer} \cite{jain2023oneformer}.

To measure the spatial alignment of the estimated pose against the ground truth, we evaluate PCK, PCKh, AKD, MAE, MSE, and SIM. Additionally, we compute the IOU between the predicted and target pose bounding boxes to measure the correctness of inferred scales. Table \ref{tab:ablation} summarizes the evaluation scores for every network variant in our ablation study. The analysis shows that additional supervision from other modalities generally results in better performance. Introducing the MCMA block into the architecture further improves this performance gain by a significant margin, reflecting the efficacy of the proposed approach for robust scene context representation. Interestingly, we observe that \emph{semantic segmentation maps} generally perform better than \emph{depth maps} as auxiliary contextual input. We hypothesize that the object labels in a segmentation map provide additional semantic context to the model. Fig. \ref{fig:ablation} shows a visual comparison of the predicted pose from \emph{depth}-context against \emph{semantic}-context.

\textcolor{black}{To investigate the impact of semantic label granularity, pose templates, and dedicated VAEs for scale and deformation on the proposed method, we analyze six additional configurations \texttt{\textbf{A - F}} of the architecture. The first four models \texttt{\textbf{A - D}} use different numbers of semantic labels in the segmentation maps. Specifically, configuration \texttt{\textbf{A}} uses 2 labels for \emph{foreground} (all objects merged) and \emph{background}. Configuration \texttt{\textbf{B}} uses 3 labels by dividing the foreground objects into \emph{non-human objects} and \emph{humans}. Configuration \texttt{\textbf{C}} uses 4 labels by further splitting the non-human object labels into \emph{fixed} (wall, floor, stairs) and \emph{movable} (table, chair, bed) object categories. Configuration \texttt{\textbf{D}} retains all 150 initially estimated semantic labels unaltered. To verify the requirements of multiple pose templates, Configuration \texttt{\textbf{E}} drops the classifier from the architecture and uses the first template as a fixed predefined pose. To validate the necessity of 2 dedicated VAEs for separately estimating scale and deformation parameters, Configuration \texttt{\textbf{F}} uses a single unified VAE to predict both scale and deformation parameters as a single vector. Finally, Configuration \texttt{\textbf{G}} denotes the proposed architecture that uses 8 semantic labels, 30 pose templates with a template classifier, and 2 dedicated VAEs for estimating scale and deformation parameters separately.}

\begin{table}[t]
\centering
\caption{\textcolor{black}{Quantitative ablation analysis of different variations of the proposed network architecture.}}
\label{tab:ablation_2}
\resizebox{\linewidth}{!}{%
\begin{tabular}{ccccccccc}
\hline
\rowcolor[HTML]{ECECEC}
  \textcolor{black}{\textbf{Model}}            &
  \textcolor{black}{\textbf{PCK $\uparrow$}}   &
  \textcolor{black}{\textbf{PCKh $\uparrow$}}  &
  \textcolor{black}{\textbf{AKD $\downarrow$}} &
  \textcolor{black}{\textbf{MAE $\downarrow$}} &
  \textcolor{black}{\textbf{MSE $\downarrow$}} &
  \textcolor{black}{\textbf{SIM $\uparrow$}}   &
  \textcolor{black}{\textbf{IOU $\uparrow$}}   \\ \hline
\textcolor{black}{\texttt{\textbf{A}}} &
  \textcolor{black}{0.217}  &
  \textcolor{black}{0.267}  &
  \textcolor{black}{11.941} &
  \textcolor{black}{7.223}  &
  \textcolor{black}{52.88}  &
  \textcolor{black}{0.9821} &
  \textcolor{black}{0.409}  \\
\textcolor{black}{\texttt{\textbf{B}}} &
  \textcolor{black}{0.292}  &
  \textcolor{black}{0.351}  &
  \textcolor{black}{~9.112} &
  \textcolor{black}{5.917}  &
  \textcolor{black}{45.11}  &
  \textcolor{black}{0.9901} &
  \textcolor{black}{0.477}  \\
\textcolor{black}{\texttt{\textbf{C}}} &
  \textcolor{black}{0.377}  &
  \textcolor{black}{0.441}  &
  \textcolor{black}{~7.019} &
  \textcolor{black}{4.508}  &
  \textcolor{black}{37.79}  &
  \textcolor{black}{0.9964} &
  \textcolor{black}{0.564}  \\
\textcolor{black}{\texttt{\textbf{D}}} &
  \textcolor{black}{0.376}  &
  \textcolor{black}{0.436}  &
  \textcolor{black}{~7.313} &
  \textcolor{black}{4.573}  &
  \textcolor{black}{40.12}  &
  \textcolor{black}{0.9959} &
  \textcolor{black}{0.557}  \\
\textcolor{black}{\texttt{\textbf{E}}} &
  \textcolor{black}{0.352}  &
  \textcolor{black}{0.400}  &
  \textcolor{black}{~7.915} &
  \textcolor{black}{4.841}  &
  \textcolor{black}{47.69}  &
  \textcolor{black}{0.9951} &
  \textcolor{black}{0.541}  \\
\textcolor{black}{\texttt{\textbf{F}}} &
  \textcolor{black}{0.369}  &
  \textcolor{black}{0.423}  &
  \textcolor{black}{~7.206} &
  \textcolor{black}{4.523}  &
  \textcolor{black}{38.02}  &
  \textcolor{black}{0.9961} &
  \textcolor{black}{0.563}  \\
\rowcolor[HTML]{CCFFCC}
\textcolor{black}{\texttt{\textbf{G}}} &
  \textcolor{black}{\textbf{0.433}}  &
  \textcolor{black}{\textbf{0.503}}  &
  \textcolor{black}{\textbf{~6.352}} &
  \textcolor{black}{\textbf{3.972}}  &
  \textcolor{black}{\textbf{29.81}}  &
  \textcolor{black}{\textbf{0.9969}} &
  \textcolor{black}{\textbf{0.566}}  \\ \hline
\end{tabular}%
}
\end{table}

\textcolor{black}{Table \ref{tab:ablation_2} summarizes evaluation scores for all the network configurations. The results show that using \emph{too few} or \emph{too many} semantic labels does not contribute towards performance improvements. Also, with a fixed template, the linear deviations between the target and template keypoints can vary more randomly. For example, the deviation of a \emph{standing} pose template from a \emph{standing} target pose is often much smaller than from a \emph{sitting} target pose. However, with multiple pose templates, the model estimates a probable pose (template) first and then samples linear deformation parameters from a more predictable range to translate the template into the target pose. Likewise, the possible ranges of scaling and deformation parameters are widely different. The scaling parameters are the height and width of the minimal bounding box around a human pose. These values are much larger than the deformation parameters comprising small linear deviations between two sets of pose keypoints. So, forcing the architecture to infer the scaling and deformation parameters with a single unified VAE causes noticeable instability due to poor normalization. These observations further justify the proposed network design.}

\subsection{Limitations}
\label{sec:experiments_limitations}

\begin{figure}[t]
\centering
\captionsetup[subfloat]{labelfont=bf}
\subfloat{\includegraphics[width=0.32\linewidth]{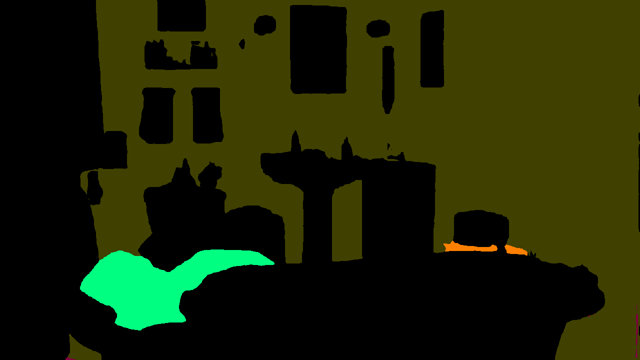}}\hfil
\subfloat{\includegraphics[width=0.32\linewidth]{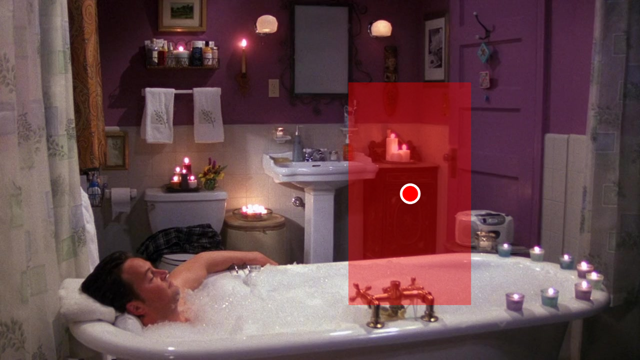}}\hfil
\subfloat{\includegraphics[width=0.32\linewidth]{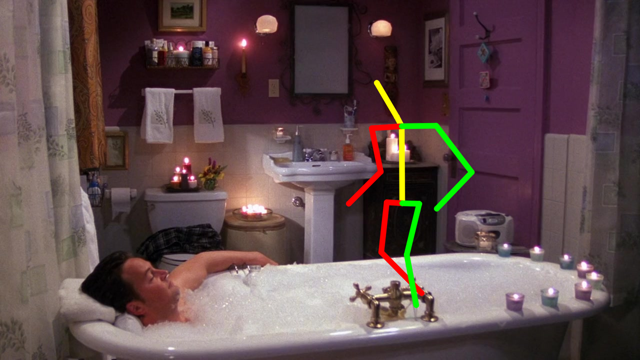}}
\vspace{-0.75em}
\subfloat{\includegraphics[width=0.32\linewidth]{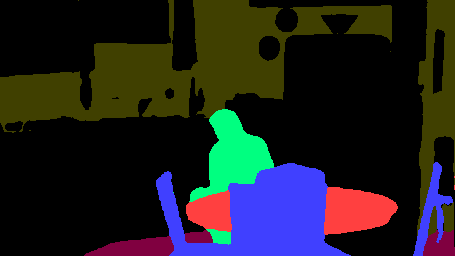}}\hfil
\subfloat{\includegraphics[width=0.32\linewidth]{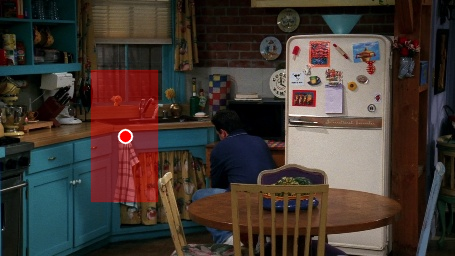}}\hfil
\subfloat{\includegraphics[width=0.32\linewidth]{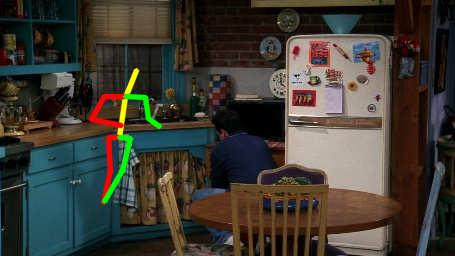}}
\vspace{-0.75em}
\subfloat{\includegraphics[width=0.32\linewidth]{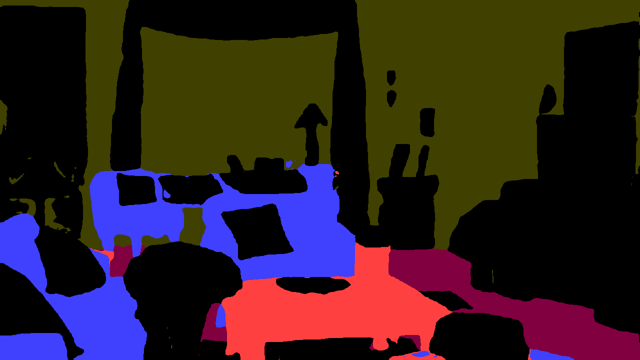}}\hfil
\subfloat{\includegraphics[width=0.32\linewidth]{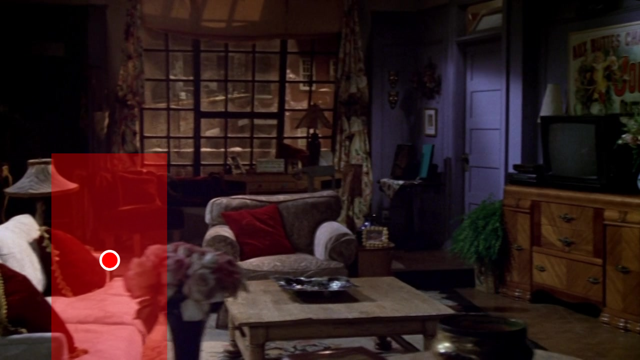}}\hfil
\subfloat{\includegraphics[width=0.32\linewidth]{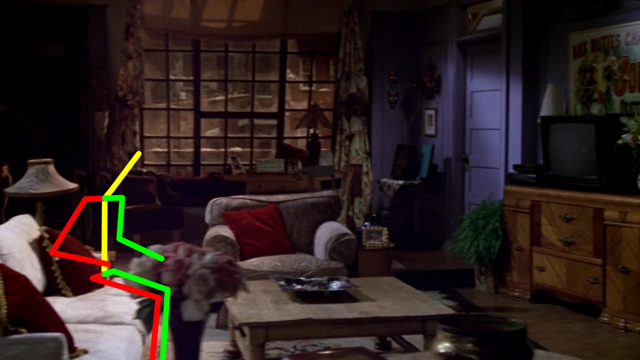}}
\caption{Visual examples of limiting cases inferred from the proposed method. \textbf{(Left)} Auxiliary semantic context. \textbf{(Middle)} Estimated bounding region. \textbf{(Right)} Estimated human pose.}
\label{fig:limitations}
\end{figure}

Estimating a valid pose for a non-existent person in complex scenes is a fundamentally challenging problem with multiple acceptable solutions other than the ground truth. For example, for a scene containing a bed, the probable pose can be standing, sitting or lying down, with many feasible variations for each case. While on most occasions, the proposed method generally infers realistic poses with affordance-aware human interactions, there are a few instances when the technique fails to sample an acceptable posture. Alongside the advantages of the modularity and flexibility in the disentangled multi-stage approach, a potential error propagation problem also exists. More precisely, inferential error in an earlier stage may propagate through later stages, negatively impacting the overall predictive performance of the pipeline. We show a few examples of such limiting cases in Fig. \ref{fig:limitations}, illustrating the misinterpretation in sampled location \textbf{(top row)}, scale \textbf{(middle row)} or deformation \textbf{(bottom row)} by the proposed method.

\subsection{Downstream applications}
\label{sec:experiments_applications}

\begin{figure}[t]
\centering
\captionsetup[subfloat]{labelfont=bf}
\subfloat{\includegraphics[width=0.32\linewidth]{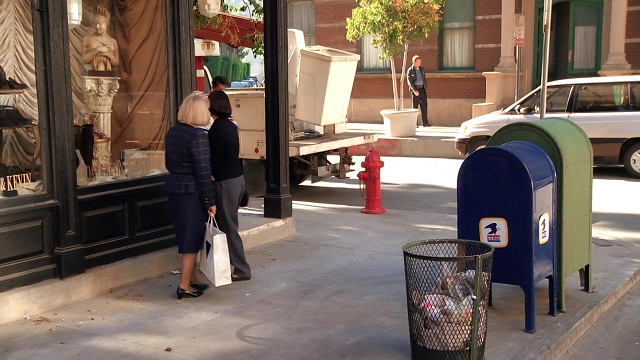}}\hfil
\subfloat{\includegraphics[width=0.32\linewidth]{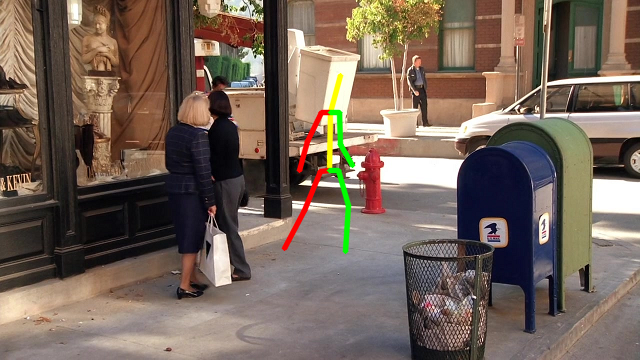}}\hfil
\subfloat{\includegraphics[width=0.32\linewidth]{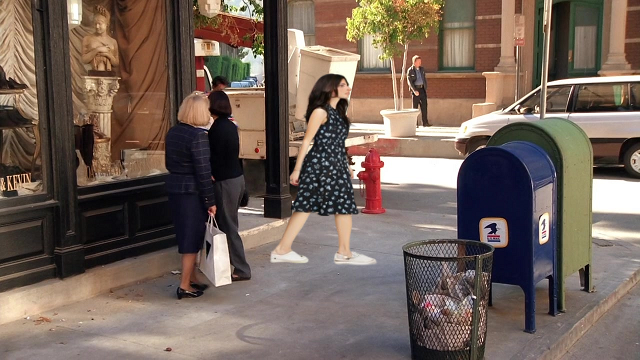}}
\vspace{-0.75em}
\subfloat{\includegraphics[width=0.32\linewidth]{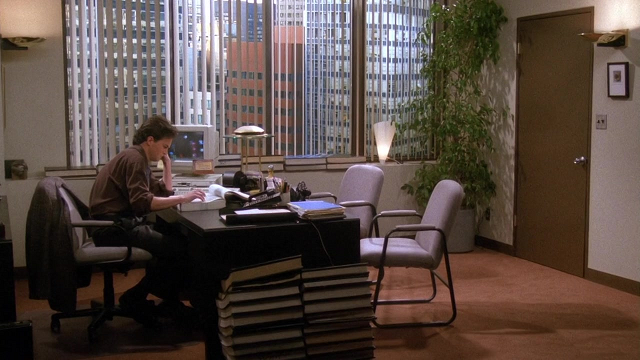}}\hfil
\subfloat{\includegraphics[width=0.32\linewidth]{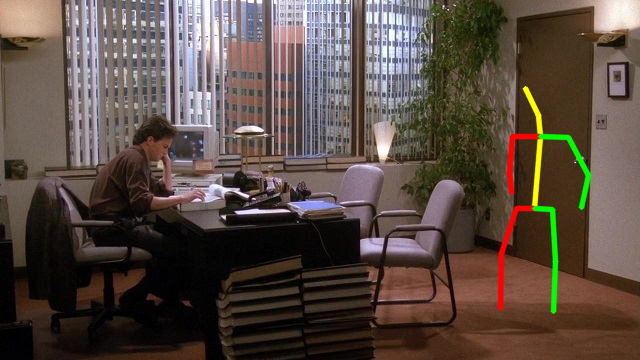}}\hfil
\subfloat{\includegraphics[width=0.32\linewidth]{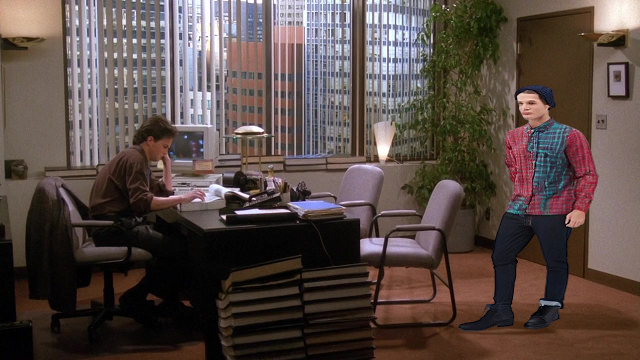}}
\vspace{-0.75em}
\subfloat{\includegraphics[width=0.32\linewidth]{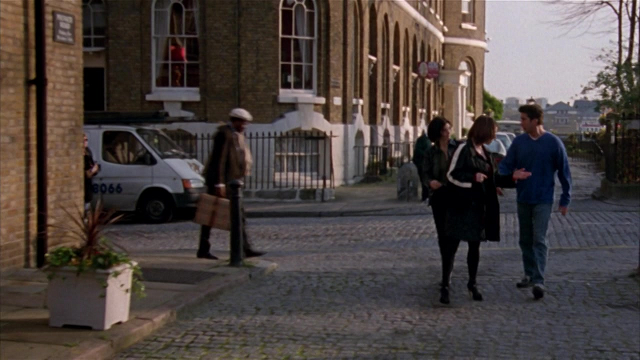}}\hfil
\subfloat{\includegraphics[width=0.32\linewidth]{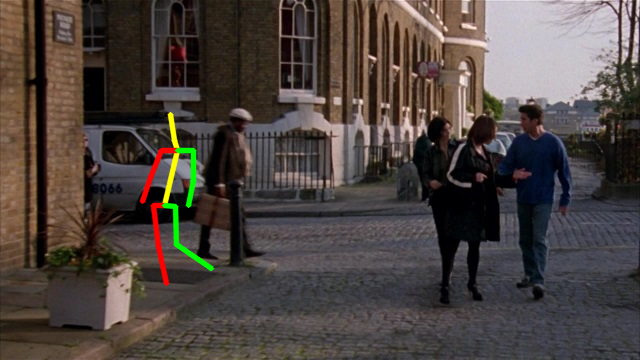}}\hfil
\subfloat{\includegraphics[width=0.32\linewidth]{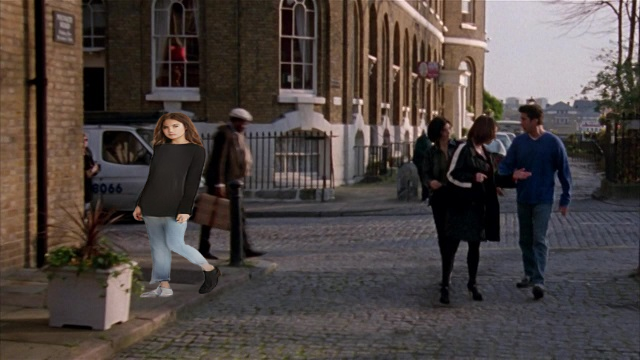}}
\caption{Visual examples of downstream rendering of human instances. \textbf{(Left)} Input scene. \textbf{(Middle)} Estimated pose by the proposed method. \textbf{(Right)} Rendered person using PIDM \cite{bhunia2023person}.}
\label{fig:rendering}
\end{figure}

The ability to sample semantically-valid scene-aware complex human poses directly facilitates downstream tasks such as novel person instance generation using off-the-shelf pose transfer or pose rendering techniques. Such downstream rendering to inject novel person instances into complex scenes is critical in various application domains, including augmented / virtual reality, digital media, and synthetic data generation. While the state-of-the-art keypoint-based person generation techniques provide high-quality photorealistic rendering, the algorithms also demand precise supervision of the target pose. Therefore, the proposed method must sample a valid and accurate human pose for successful downstream rendering. Fig. \ref{fig:rendering} shows a few examples of rendered persons using an off-the-shelf pose transfer technique \emph{PIDM} \cite{bhunia2023person} on sampled poses from our method, demonstrating the geometric correctness of the predicted keypoints.

\section{Conclusions}
\label{sec:conclusions}

This work investigates the fundamental computer vision problem of semantically constrained human affordance generation in complex environments. We discuss the main challenges of estimating realistic poses for non-existent persons by exploring current approaches in the literature to address the problem. The vast majority of the existing techniques are built around architectural innovations of the network without any significant emphasis on the semantic understanding of the scene. By introducing a novel cross-attention mechanism, we show that a robust semantic representation of the scene context can improve the generation performance by a significant margin. In addition, our approach provides a fully automated inference pipeline. We demonstrate the efficacy of the proposed method with visual results, quantitative analyses on pose alignment, and an opinion-based subjective user evaluation.

\vspace{1.0em}


\bibliographystyle{IEEEtran}
\bibliography{IEEEabrv,main}









\vfill

\end{document}